\documentclass{pnastwo}

\usepackage{amssymb,amsfonts,amsmath}
\usepackage{verbatim}
\usepackage{varwidth}

\usepackage[svgnames]{xcolor} 
\usepackage{graphicx}
\usepackage[margin= 0.8in]{geometry}

\usepackage{algorithm,algpseudocode,lipsum}
\usepackage{caption}
\usepackage{framed}
\floatname{algorithm}{Procedure}

\newcommand{\reals}{\mathbb{R}}

\newcommand{\im}{I}

\newcommand{\XX}{\mathcal X} 
 
\newcommand{\R}{\mathbb R}
\providecommand{\nor}[1]{\left\lVert {#1} \right\rVert}
\providecommand{\scal}[2]{\left\langle{#1},{#2}\right\rangle}

\newcommand*{\titleAT}{\begingroup 
\newlength{\drop} 
\drop=0.05\textheight 

\includegraphics[scale=1.5]{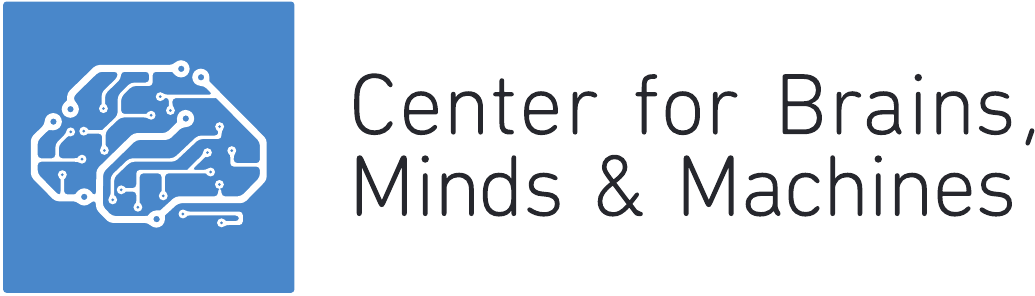}

\textcolor{CornflowerBlue}{\rule{\textwidth}{3 pt}}\par 
\vspace{2pt}\vspace{-\baselineskip} 
\rule{\textwidth}{0.4pt}\par 

\vspace{\drop} 
\textbf{\large{CBMM Memo No. \memonumber}}\quad \quad \quad\quad \quad \quad \quad\quad\quad \quad\quad\quad      \textbf{\large{\memodate}}

\vspace{\drop}
\begin{center}
\textbf{\huge{\memotitle}}\\
\vspace{0.4\drop}
\textbf{\Large{by}}\\
\vspace{0.4\drop}
\textbf{\large{\memoauthors}}
\end{center}
\vspace{\drop}
\textbf{\large{\noindent Abstract}:} {\memoabstract}

\vfill 

\textcolor{CornflowerBlue}{\rule{\textwidth}{3 pt}}\par 
\vspace{2pt}\vspace{-\baselineskip} 
\rule{\textwidth}{0.4pt}\par

\begin{minipage}{.15\linewidth}
\includegraphics[scale=0.1]{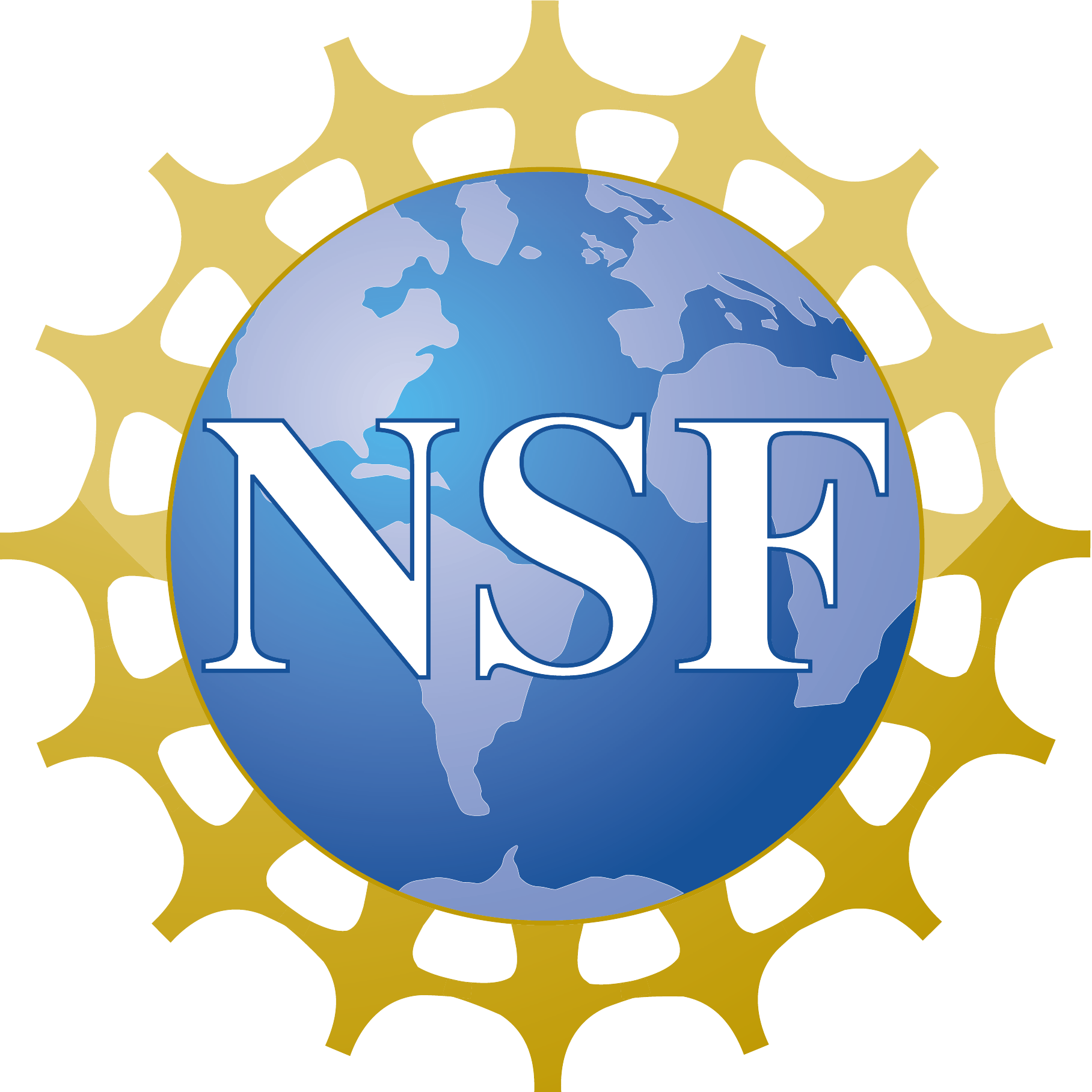}
\end{minipage}
\begin{minipage}{.84\linewidth}
\textbf{\large{This work was supported by the Center for Brains, Minds and Machines (CBMM), funded by NSF STC award  CCF - 1231216.}}
\end{minipage}
\endgroup}

\newtheorem{definition*}{Definition}

\url{http://cbmm.mit.edu}
\copyrightyear{2014}
\issuedate{March, 2014}
\volume{}
\issuenumber{}

\begin{document}

\def\memonumber{001} 
\def\memodate{\today} 
\def\memotitle{Unsupervised learning of invariant representations with low sample complexity: the magic of sensory cortex or a new framework for machine learning?} 
\def\memoauthors{ Fabio Anselmi, Joel Z. Leibo, Lorenzo Rosasco, Jim Mutch, Andrea Tacchetti and Tomaso Poggio.}
\def\memoabstract{The present phase of Machine Learning is characterized by supervised learning algorithms relying on large sets of labeled examples ($n \to \infty$). The next phase is likely to focus on algorithms capable of learning from very few labeled examples ($n \to 1$), like humans seem able to do. We propose an approach to this problem and describe the underlying theory, based on the unsupervised, automatic learning of a ``good'' representation for supervised learning, characterized by small sample complexity ($n$). We consider the case of visual object recognition though the theory applies to other domains. The starting point is the conjecture, proved in specific cases, that image representations which are invariant to translations, scaling and other transformations can considerably reduce the sample complexity of learning. We prove that an invariant and unique (discriminative) signature can be computed for each image patch, $I$, in terms of empirical distributions of the dot-products between $I$ and a set of templates stored during unsupervised learning. A module performing filtering and pooling, like the simple and complex cells described by Hubel and Wiesel, can compute such estimates. Hierarchical architectures consisting of this basic Hubel-Wiesel moduli inherit its properties of invariance, stability, and discriminability while capturing the compositional organization of the visual world in terms of wholes and parts. The theory extends existing deep learning convolutional architectures for image and speech recognition. It also suggests that the main computational goal of the ventral stream of visual cortex is to provide a hierarchical representation of new objects/images which is invariant to transformations, stable, and discriminative for recognition---and that this representation may be continuously learned in an unsupervised way during development and visual experience.}

\titleAT 

\title{Unsupervised learning of invariant representations with low sample complexity: the magic of sensory cortex or a new framework for machine learning? }

\author{Fabio Anselmi \affil{1}{Center for Brains, Minds and Machines, Massachusetts Institute of Technology, Cambridge, MA 02139}\affil{2}{Istituto Italiano di Tecnologia, Genova, 16163}, Joel Z. Leibo \affil{1}{}, Lorenzo Rosasco \affil{1}{}\affil{2}{}, Jim Mutch \affil{1}{}, Andrea Tacchetti \affil{1}{}\affil{2}{}, \and Tomaso Poggio \affil{1}{}\affil{2}{}}

\maketitle

\begin{article}
\begin{abstract}
The present phase of Machine Learning is characterized by supervised learning algorithms relying on large sets of labeled examples ($n \to \infty$). The next phase is likely to focus on algorithms capable of learning from very few labeled examples ($n \to 1$), like humans seem able to do. We propose an approach to this problem and describe the underlying theory, based on the unsupervised, automatic learning of a ``good'' representation for supervised learning, characterized by small sample complexity ($n$). We consider the case of visual object recognition though the theory applies to other domains. The starting point is the conjecture, proved in specific cases, that image representations which are invariant to translations, scaling and other transformations can considerably reduce the sample complexity of learning. We prove that an invariant and unique (discriminative) signature can be computed for each image patch, $I$, in terms of empirical distributions of the dot-products between $I$ and a set of templates stored during unsupervised learning. A module performing filtering and pooling, like the simple and complex cells described by Hubel and Wiesel, can compute such estimates. Hierarchical architectures consisting of this basic Hubel-Wiesel moduli inherit its properties of invariance, stability, and discriminability while capturing the compositional organization of the visual world in terms of wholes and parts. The theory extends existing deep learning convolutional architectures for image and speech recognition. It also suggests that the main computational goal of the ventral stream of visual cortex is to provide a hierarchical representation of new objects/images which is invariant to transformations, stable, and discriminative for recognition---and that this representation may be continuously learned in an unsupervised way during development and visual experience.\footnote{\small {\bf Notes on versions and dates} The current paper evolved from one that first appeared online
in Nature Precedings on July 20, 2011 (npre.2011.6117.1). It follows a CSAIL technical report which appeared on December 30th,
2012,MIT-CSAIL-TR-2012-035 and a CBCL paper, Massachusetts Institute of Technology, Cambridge, MA, April 1, 2013 by the title  "Magic Materials: a theory of deep hierarchical architectures for learning sensory representations"(\cite{MM2013}). Shorter papers describing isolated
aspects of the theory have also appeared:\cite{Leibo2011b, Liao2013}.}
\end{abstract}

\keywords{Invariance|Hierarchy|Convolutional networks|Visual cortex}

\begin{figure}\centering
\includegraphics[width=0.40\textwidth]{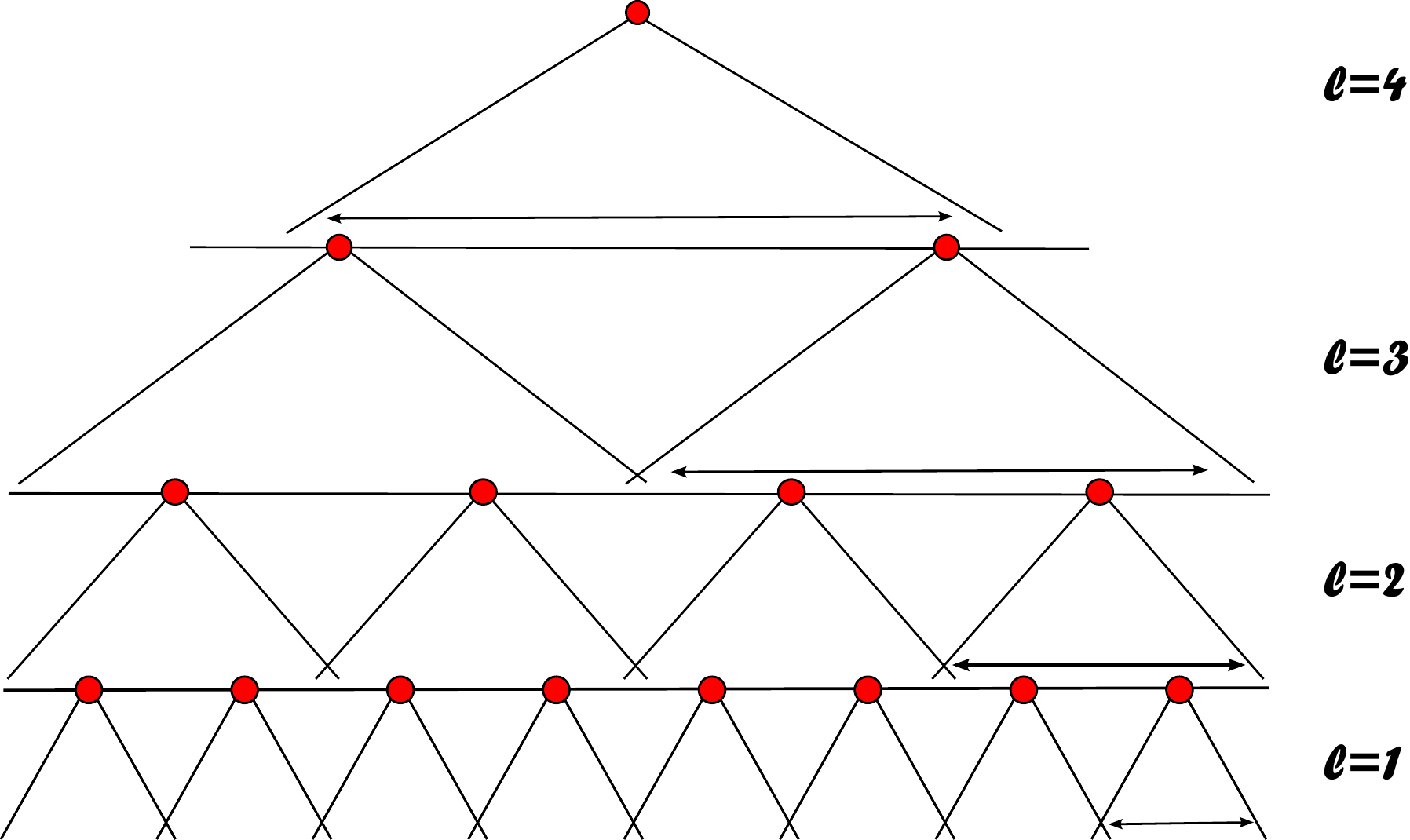}
\caption{A hierarchical architecture built from HW-modules. Each red circle represents the signature vector computed by the associated module (the outputs of complex cells) and  double arrows represent its receptive fields -- the part of the (neural) image visible to the module (for translations this is also the pooling range). The ``image'' is at level $0$, at the bottom. The vector computed at the top of the hierarchy consists of invariant features for the whole image and is usually fed as input to a supervised learning machine such as a classifier; in addition signatures from modules at intermediate layers may also be inputs to classifiers for objects and parts.\label{Wedge}}
\end{figure}

It is known that Hubel and Wiesel's original proposal \cite{Hubel1962}  for visual area V1---of a module consisting of complex cells (C-units) combining the outputs of sets of simple cells (S-units) with identical orientation preferences but differing retinal positions---can be used to construct translation-invariant detectors. This is the insight underlying many networks for visual recognition, including HMAX \cite{Riesenhuber1999a} and convolutional neural nets \cite{Fukushima1980, LeCun1989}. We show here how the original idea can be expanded into a comprehensive theory of visual recognition relevant for computer vision and possibly for visual cortex. The first step in the theory is the conjecture that a representation of images and image patches, with a feature vector that is invariant to a broad range of transformations---such as translation, scale, expression of a face, pose of a body, and viewpoint---makes it possible to recognize objects from only a few labeled examples, as humans do. The second step is proving that hierarchical architectures of Hubel-Wiesel (`HW') modules (indicated by $\bigwedge$ in Fig. \ref{Wedge}) can provide such invariant representations while maintaining discriminative information about the original image.
Each $\bigwedge$-module provides a feature vector, which we call a {\it signature},  for the part of the visual field that is inside its ``receptive field''; the signature is invariant to  ($\reals^{2}$) affine transformations within the receptive field.  The hierarchical architecture, since it computes a set of signatures for different parts of the image,  is proven to be invariant to the rather general family of locally affine transformations (which includes globally affine transformations of the whole image).
The basic HW-module is at the core of the properties of the architecture. This paper focuses first on its characterization and then outlines the rest of the theory, including its connections with machine learning, machine vision and neuroscience. Most of the theorems are in the supplementary information, where in the interest of telling a complete story we quote some results which are described more fully elsewhere \cite{MM2013,Leibo2011b, Liao2013}.

\section{Invariant representations and sample complexity}
One could argue that the most important aspect of intelligence is the ability to learn. How do present supervised learning algorithms compare with brains?  One of the most obvious differences is the ability of people and animals to learn from very few labeled examples.  A child, or a monkey, can learn a recognition task from just a few examples. The main motivation of this paper is the conjecture that the key to reducing the sample complexity of object recognition is invariance to  transformations. Images of the same object usually differ from each other because of simple transformations such as  translation, scale (distance) or more complex deformations such as viewpoint (rotation in depth) or change in pose (of a body) or expression (of a face).

The conjecture is supported by previous theoretical work showing that \emph{almost all the complexity} in recognition tasks is often due to the viewpoint and illumination nuisances that swamp the intrinsic characteristics of the object \cite{lee2012video}. It implies that in many cases, recognition---i.e., both identification, e.g., of a specific car relative to other cars---as well as categorization, e.g., distinguishing between cars and airplanes---would be much easier (only a small number of training examples would be needed to achieve a given level of performance, i.e. $n\rightarrow 1$), {\it if} the images of objects were rectified with respect to all transformations, or equivalently, if the image representation itself were invariant. In SI Appendix, section 0 we provide a proof of the conjecture for the special case of translation (and for obvious generalizations of it).

\begin{figure}
\begin{center}
\includegraphics[scale=.45]{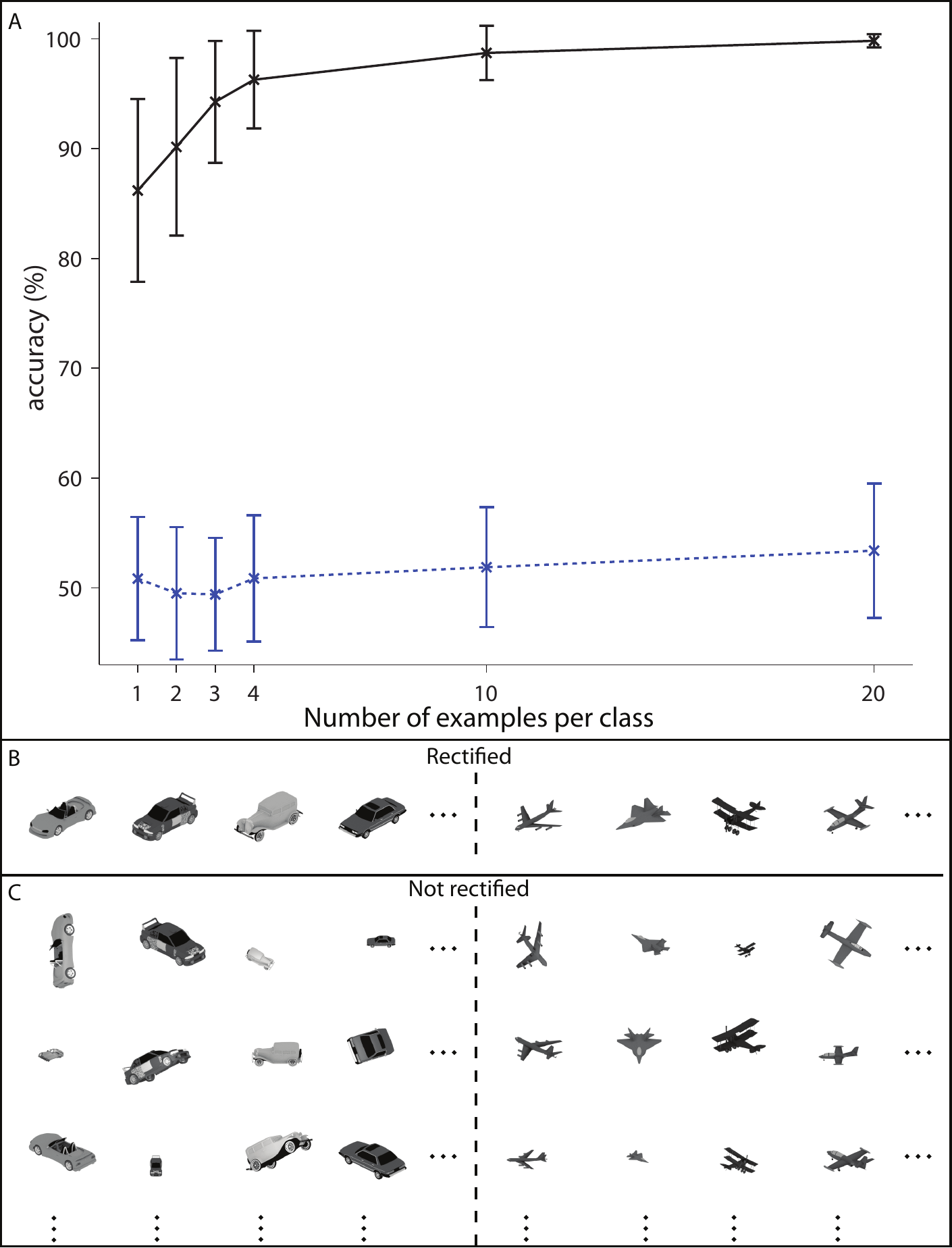}
\end{center}
\caption{Sample complexity for the task of categorizing cars vs airplanes from their raw pixel representations (no preprocessing). A. Performance of a nearest-neighbor classifier (distance metric = 1 - correlation) as a function of the number of examples per class used for training. Each test used 74 randomly chosen images to evaluate the classifier. Error bars represent +/- 1 standard deviation computed over 100 training/testing splits using different images out of the full set of 440 objects $\times$ number of transformation conditions. Solid line: The rectified task. Classifier performance for the case where all training and test images are rectified with respect to all transformations; example images shown in B. Dashed line: The unrectified task. Classifier performance for the case where variation in position, scale, direction of illumination, and rotation around any axis (including rotation in depth) is allowed; example images shown in C. The images were created using 3D models from the Digimation model bank and rendered with Blender.\label{fig:airplanes_vs_cars}}
\end{figure}

The case of identification is obvious since the difficulty in recognizing exactly the same object, e.g., an individual face, is only due to transformations.  In the case of categorization, consider the suggestive evidence from the classification task in Fig. \ref{fig:airplanes_vs_cars}. The figure shows that if an oracle  factors out all transformations in images of many different cars and airplanes, providing ``rectified'' images with respect to viewpoint, illumination, position and scale, the problem of categorizing cars vs airplanes becomes easy: it can be done accurately with very few labeled examples.  In this case, good performance was obtained from a single training image of each class, using a simple classifier. In other words, the sample complexity of the problem seems to be very low. We propose that the ventral stream  in visual cortex tries to approximate such an oracle, providing a quasi-invariant signature for images and image patches.

\section{Invariance and uniqueness}
Consider the problem of recognizing an image, or an image patch, independently of whether it has been transformed by the action of a group like the affine group in $\reals^2$.  We would like to associate to each object/image $I$ a \textit{signature}, i.e., a vector which is \textit{unique} and \textit{invariant} with respect to a group of transformations, $G$.  (Note that our analysis, as we will see later, is not restricted to the case of groups.)  In the following, we will consider groups that are compact and, for simplicity, finite (of cardinality $|G|$).  We indicate, with slight abuse of notation, a generic group element and its (unitary) representation with the same symbol $g$, and its action on an image as $gI(x) = I(g^{-1}x)$ (e.g., a translation, $g_{ {\xi}}I( {x}) = I( {x}- {\xi})$).  A natural mathematical object to consider is the \textit{orbit} $O_I$---the set of images $gI$ generated from a single image $I$ under the action of the group. We say that two images are equivalent when they belong to the same orbit: $I \sim I^\prime$ if $~\exists g\in G$ such that $I^\prime = gI$.  This equivalence relation formalizes the idea that an orbit is invariant and unique. Indeed, if two orbits have a point in common they are identical everywhere.  Conversely, two orbits are different if none of the images in one orbit coincide with any image in the other\cite{mirbach1994}.

How can two orbits be characterized and compared? There are several possible approaches. A distance between orbits can be defined in terms of a metric on images, but its computation is not obvious (especially by neurons). We follow here a different strategy: intuitively two empirical orbits are the same irrespective of the ordering of their points. This suggests that we consider the probability distribution $P_I$ induced by the group's action on images $I$ ($gI$ can be seen as a realization of a random variable). It is possible to prove (see Theorem 2 in  SI Appendix section 2) that if two orbits coincide then their associated distributions under the group $G$ are identical, that is
\begin{equation}
 I\sim I' \iff O_I = O_{I'} \iff P_{I}=P_{I'}.
\end{equation}
The distribution $P_I$ is thus invariant and discriminative, but it also inhabits a high-dimensional space and is therefore difficult to estimate. In particular, it is unclear how neurons or neuron-like elements could estimate it.

As argued later, neurons can effectively implement (high-dimensional) inner products, $\scal{\cdot}{\cdot}$, between inputs and stored ``templates'' which are neural images. It turns out that classical results (such as the Cramer-Wold theorem \cite{CramerWold1936}, see Theorem 3 and 4 in section 2 of SI Appendix) ensure that a  probability distribution $P_I$ can be almost uniquely characterized by $K$ one-dimensional probability distributions $P_{\scal{I}{t^k}}$ induced by the (one-dimensional) results of projections ${\scal{I}{t^k}}$, where $t^{k},\;k=1,...,K $ are a set of randomly chosen images called templates.
A probability function in $d$ variables (the image dimensionality) induces a unique set of 1-D projections which is discriminative; empirically a small number of projections is usually sufficient to discriminate among a finite number of different probability distributions.  Theorem 4 in SI Appendix section 2 says (informally) that an approximately invariant and unique signature of an image $I$ can be obtained from the estimates of $K$ 1-D probability distributions $P_{\scal{I}{t^k}}$ for $k=1,\cdots,K$. The number $K$ of projections needed to discriminate $n$ orbits, induced by $n$ images, up to precision $\epsilon$ (and with confidence $1-\delta^2$) is $K \ge \frac{2}{c\epsilon^2} \log {\frac{n}{\delta}}$, where $c$ is a universal constant.

Thus the discriminability question can be answered positively (up to $\epsilon$) in terms of empirical estimates of the one-dimensional distributions $P_{\scal{I}{t^k}}$ of projections of the image onto a finite number of templates
$t^k, ~ k=1,...,K$ under the action of the group.

\section{Memory-based learning of invariance}
Notice that the estimation of $P_{\scal{I}{t^k}}$ requires the observation of the image {\it and} ``all'' its transforms $gI$. Ideally, however, we would like to compute an invariant signature for a new object seen only once (e.g., we can recognize a new face at different distances after just one observation, i.e. $n\rightarrow 1$). It is remarkable and almost magical that this is also made possible by the projection step. The key is the observation that $\scal{gI}{t^k} = \scal{I}{g^{-1}t^{k}}$. The same one-dimensional distribution is obtained from the projections of the image and all its transformations onto a fixed template, as from the projections of the image onto all the transformations of the same template.  Indeed, the distributions of the variables $\scal{I}{g^{-1} t^k}$ and $\scal{gI}{t^k}$ are the same.  Thus it is possible for the system to store for each template $t^k$ all its transformations $g t^k$ for all $g \in G$ and later obtain an invariant signature for new images without any explicit knowledge of the transformations $g$ or of the group to which they belong.  {\it Implicit knowledge of the transformations}, in the form of the stored templates, allows the system to be {\it automatically invariant to those transformations for new inputs (see eq. $[8]$  in SI Appendix).}

Estimates of the one-dimensional probability density functions (PDFs) $P_{\scal{I}{t^k}}$ can be written in terms of histograms as $\mu_n^k(I) =1/|G| \sum_{i=1}^{|G|} \eta_n (\scal{I}{g_i t^k})$, where $\eta_{n},\;n=1,\cdots,N$ is a set of nonlinear functions (see remark 1 in SI Appendix section 1 or Theorem 6 in section 2 but also \cite{McCulloch1947}). A visual system  need not recover the actual probabilities from the empirical estimate in order to compute a unique signature. The set of $\mu_n^k(I)$ values is sufficient, since it identifies the associated orbit (see box 1 in SI Appendix). Crucially, mechanisms capable of computing invariant representations under affine transformations for future objects can be learned and maintained in an unsupervised, automatic way by storing and updating sets of transformed templates which are \emph{unrelated to those future objects}.

\section{A theory of pooling}
The arguments above make a few predictions. They require an effective normalization of the elements of the inner product (e.g. $\scal{I}{g_i t^k} \mapsto \frac{\scal{I}{g_i t^k} }{\|I\|\|g_i t^k\|}$) for the property $\scal{gI}{t^k} = \scal{I}{g^{-1}t^{k}}$ to be valid (see remark 8 of SI Appendix section 1 for the affine transformations case). Notice that invariant signatures can be computed in several ways from one-dimensional probability distributions. Instead of the $\mu_n^k(I)$ components directly representing the empirical distribution, the moments $m^k_n(I) =1/|G| \sum_{i=1}^{|G|} (\scal{I}{g_i t^k} )^n$ of the same distribution can be used \cite{koloydenko2008} (this corresponds to the choice $\eta_{n}(\cdot)\equiv (\cdot)^{n}$ ). Under weak conditions, the set of \emph{all} moments uniquely characterizes the one-dimensional distribution $P_{\scal{I}{t^k}}$ (and thus $P_I$).  $n=1$ corresponds to pooling via sum/average (and is the only pooling function that does not require a nonlinearity); $n=2$ corresponds to "energy models" of complex cells and $n=\infty$ is related to max-pooling.  In our simulations, just one of these moments usually seems to provide sufficient selectivity to a hierarchical architecture (see SI Appendix section 6). Other nonlinearities are also possible\cite{MM2013}.  The arguments of this section begin to provide a theoretical understanding of ``pooling'', giving insight into the search for the ``best'' choice in any particular setting---something which is normally done empirically \cite{Jarrett2009}. According to this theory, these different pooling functions are all invariant, each one capturing part of the full information contained in the PDFs.

\section{Implementations}
\begin{figure}\centering
\includegraphics[width=0.4\textwidth]{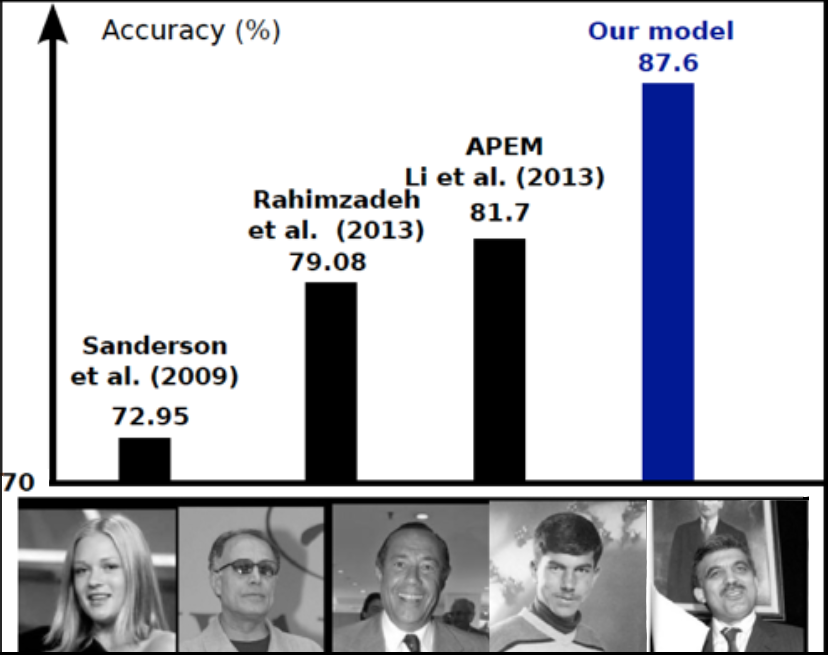}
\caption{Performance of a recent model \cite{Liao2013} (inspired by the present theory) on Labeled Faces in the Wild, a same/different person task for faces seen in different poses and in the presence of clutter.  A layer which builds invariance to translation, scaling, and limited in-plane rotation is followed by another which pools over variability induced by other transformations.\label{LFW}}
\end{figure}

The theory has strong empirical support from several specific implementations which have been shown to perform well on a number of databases of natural images. The main support is provided by HMAX, an architecture in which pooling is done with a max operation and invariance, to translation and scale, is mostly hardwired (instead of learned). Its performance on a variety of tasks is discussed in SI Appendix section 6. Good performance is also achieved by other very similar architectures \cite{Pinto2009a}.  This class of existing models inspired the present theory, and may now be seen as special cases of it.  Using the principles of invariant recognition the theory makes explicit, we have now begun to develop models that incorporate invariance to more complex transformations which cannot be solved by the architecture of the network, but must be learned from examples of objects undergoing transformations.  These include non-affine and even non-group transformations, allowed by the hierarchical extension of the theory (see below).  Performance for one such model is shown in Figure~\ref{LFW} (see caption for details).

\section{Extensions of the Theory}

\subsection{Invariance Implies Localization and Sparsity}

The core of the theory applies without qualification to compact groups such as rotations of the image in the image plane. Translation and scaling are however only locally compact, and in any case, each of the modules of Fig. \ref{Wedge} observes only a part of the transformation's full range. Each $\bigwedge$-module has a finite pooling range, corresponding to a finite ``window'' over the orbit associated with an image. {\it Exact invariance} for each module, in the case of translations or scaling transformations, is equivalent to a condition of {\it localization/sparsity} of the dot product between image and template (see Theorem 6 and Fig. 5 in section 2 of SI Appendix). In the simple case of a group parameterized by one parameter $r$ the condition is (for simplicity $I$ and $t$ have support center in zero):

\begin{equation} \label{locm}
\scal{I}{g_r t^k} = 0  \quad |r| > a.
\end{equation}

Since this condition is a form of sparsity of the generic image $I$ w.r.t. a dictionary of templates $t^k$ (under a group), this result provides a computational justification for {\it sparse} encoding in sensory cortex\cite{Olshausen1996}.

It turns out that localization yields the following surprising result (Theorem 7 and 8 in SI Appendix):  {\it optimal invariance for translation and scale implies Gabor functions as templates}. Since a frame of Gabor wavelets follows from natural requirements of completeness, this may also provide a general motivation for the Scattering Transform approach of Mallat based on wavelets \cite{MallatGroupInvariantMain}.

The same Equation \ref{locm}, if relaxed to hold approximately, that is  $\scal{I_C}{g_r t^k} \approx 0 \quad |r| > a$,  becomes a {\it sparsity condition for the class of $I_C$ w.r.t. the dictionary $t^k$ under the group $G$} when restricted to a subclass $I_C$ of similar images. This property (see SI Appendix, end of section 2), which is an extension of the  compressive sensing notion of ``incoherence'',  requires that $I$ and $t^k$ have a representation with sharply peaked correlation and autocorrelation. When the condition is satisfied, the basic HW-module equipped with such templates can provide approximate invariance to non-group transformations such as rotations in depth of a face or its changes of expression (see Proposition 9, section 2, SI Appendix).
In summary, Equation \ref{locm} can be satisfied in two different {\it regimes}. The first one, exact and valid for generic $I$, yields optimal Gabor templates. The second regime,  approximate and valid for specific subclasses of $I$,  yields highly tuned templates, specific for the subclass. Note that this argument suggests generic, Gabor-like templates in the first layers of the hierarchy and highly specific templates at higher levels.  (Note also that incoherence improves with increasing dimensionality.)

\subsection{Hierarchical architectures} 
\begin{figure}\centering
\includegraphics[width=0.4\textwidth]{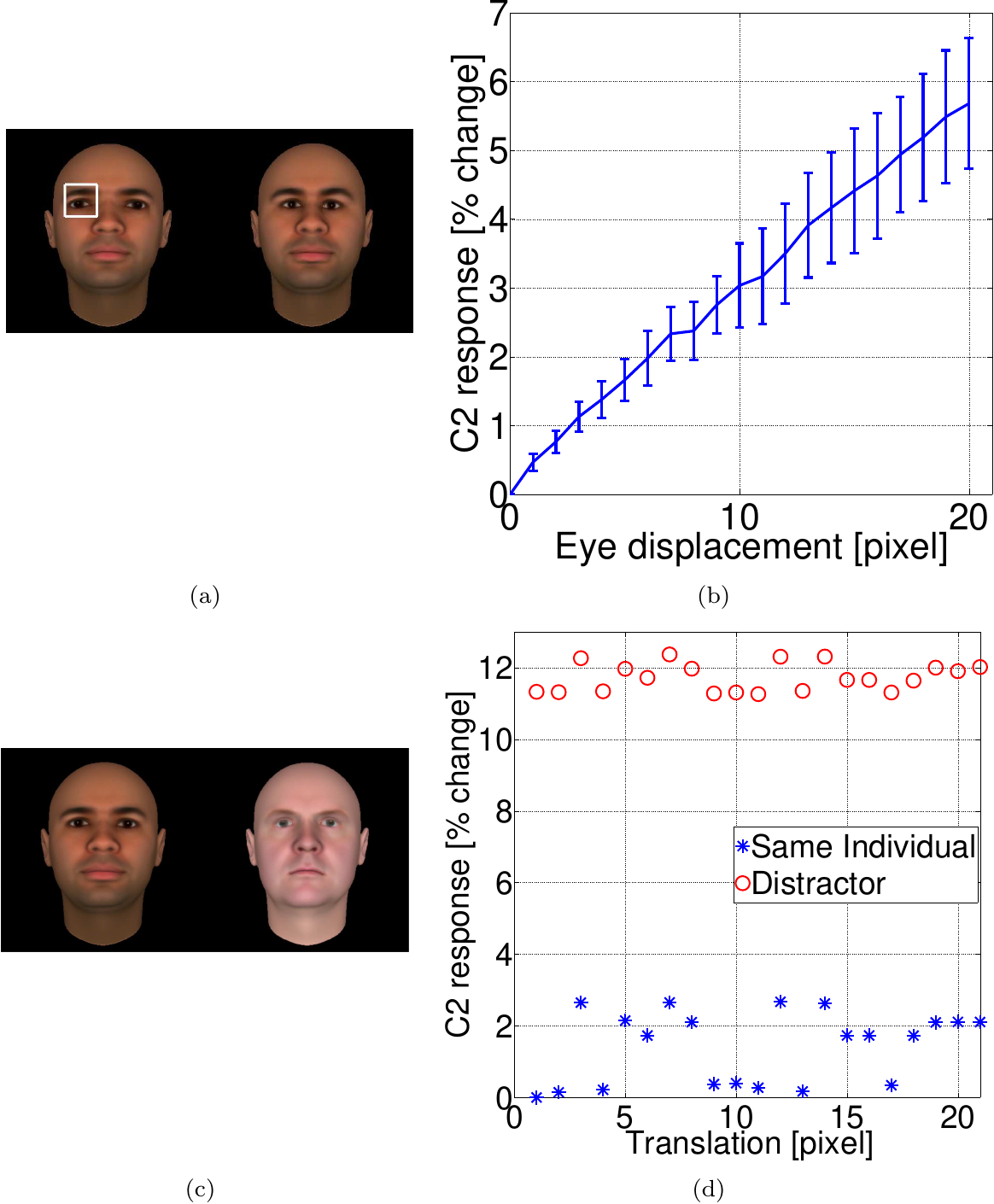}
\caption{
Empirical demonstration of the properties of invariance,
stability and uniqueness of the hierarchical architecture in a specific 2 layers implementation (HMAX). Inset (a)
shows the reference image on the left and a deformation of it (the
eyes are closer to each other) on the right; (b) shows the relative change in signature provided
by $128$ HW-modules at layer $2$ $(C_2)$ whose receptive fields contain the whole face.
This signature vector is (Lipschitz) stable with respect
to the deformation. Error bars represent $\pm 1$ standard deviation. Two different images (c) are presented
at various location in the visual field. In (d) the relative change of the signature vector for different values of translation.
The signature vector is invariant to global translation and discriminative (between the two faces). In this example the HW-module represents the top of a hierarchical, convolutional architecture. The images we used were $200\times 200$ pixels and error bars represent $\pm 1$  standard deviation.\label{invariancestability}}
\end{figure}

We have focused so far on the basic HW-module. Architectures consisting of such modules can be single-layer as well as multi-layer (hierarchical) (see Fig. \ref{Wedge}).  In our theory, the key property of hierarchical architectures of repeated HW-modules---allowing the recursive use of modules in multiple layers---is the property of {\it covariance}.  By a covariant response at layer $\ell$ we mean that
the distribution of the values of each projection is the same if we consider the image or the template transformations, i.e. (see Property 1 and Proposition 10 in section 3, SI Appendix), $distr(\scal{\mu_{\ell}(gI)}{\mu_{\ell}(t^k)})=distr(\scal{\mu_{\ell}(I)}{\mu_{\ell}(gt^k)}),\;\forall k$.\\
One-layer networks can achieve invariance to {\it global} transformations of the whole image while providing a unique global signature which is stable with respect to small perturbations of the image (see Theorem 5 in section 2 of SI Appendix and \cite{MM2013}). The two main reasons for a hierarchical architecture such as Fig. \ref{Wedge} are (a) the need to compute an invariant representation not only for the whole image but especially for all parts of it, which may contain objects and object parts, and (b) invariance to global transformations that are not affine, but are locally affine, that is, affine within the pooling range of some of the modules in the hierarchy. Of course, one could imagine local and global one-layer architectures used in the same visual system without a hierarchical configuration, but there are further reasons favoring hierarchies including compositionality and reusability of parts. In addition to the issues of sample complexity and connectivity, one-stage architectures are unable to capture the hierarchical organization of the visual world where scenes are composed of objects which are themselves composed of parts.  Objects can move in a scene relative to each other without changing their identity and often changing the scene only in a minor way; the same is often true for parts within an object.  Thus global and local signatures from all levels of the hierarchy must be able to access memory in order to enable the categorization and identification of whole scenes as well as of patches of the image corresponding to objects and their parts. Fig.  \ref{invariancestability} show examples of invariance and stability for wholes and parts. In the architecture of Fig. \ref{Wedge}, each $\bigwedge$-module provides uniqueness, invariance and stability at different levels, over increasing ranges from bottom to top. Thus, in addition to the desired properties of invariance, stability and discriminability, these architectures match the hierarchical structure of the visual world and the need to retrieve items from memory at various levels of size and complexity. The results described here are part of a general theory of hierarchical architectures which is beginning to take form (see \cite{MM2013,MallatGroupInvariantMain,Soatto2011,smale2010mathematics}) around the basic function of computing invariant representations.

The property of compositionality discussed above is related to the efficacy of hierarchical architectures vs. one-layer architectures in dealing with the problem of partial occlusion and the more difficult problem of clutter in object recognition. Hierarchical architectures are better at recognition in clutter than one-layer networks \cite{serre2005theory} because they provide signatures for image patches of several sizes and locations. However, hierarchical feedforward architectures cannot fully solve the problem of clutter. More complex (e.g.  recurrent) architectures are likely needed for human-level recognition in clutter (see for instance \cite{Chikkerur2010,George2005,geman2006invariance}) and for other aspects of human vision. It is likely that much of the circuitry of  visual cortex is required by these recurrent computations, not considered in this paper.

\section{Visual Cortex}
The theory described above effectively maps the computation of an invariant signature onto well-known capabilities of cortical neurons. A key difference between the basic elements of our digital computers and neurons is the number of connections: $3$ vs. $10^3 - 10^4$ synapses per cortical neuron. Taking into account basic properties of synapses, it follows that a single neuron can compute high-dimensional ($10^3 - 10^4$) inner products between input vectors and the stored vector of synaptic weights \cite{McCulloch1943}.\\
Consider an HW-module of ``simple'' and ``complex'' cells \cite{Hubel1962} looking at the image through a window defined by their receptive fields (see SI Appendix, section 2, POG).  Suppose that images of objects in the visual environment undergo affine transformations. During development---and more generally, during visual experience---a set of $|G|$ simple cells store in their synapses an image patch $t^k$ and its transformations $g_1t^k,...,g_{|G|}t^k$---one per simple cell.  This is done, possibly at separate times, for $K$ different image patches $t^k$ (templates), $k=1,\cdots,K$. Each $g t^k$ for $g \in G$ is a sequence of frames, literally a movie of image patch $t^k$ transforming. There is a very {\it simple, general, and powerful way to learn} such unconstrained transformations. Unsupervised (Hebbian) learning is the main mechanism: for a ``complex'' cell to pool over several simple cells, the key is an unsupervised Foldiak-type rule: {\it cells that fire together are wired together}. At the level of complex cells this rule determines {\it classes of equivalence} among simple cells -- reflecting observed {\it time correlations in the real world, that is, transformations} of the image.  Time continuity, induced by the Markovian physics of the world, allows associative labeling of stimuli based on their temporal contiguity.

Later, when an image is presented, the simple cells compute $\scal{I}{ g_{i}t^k}$ for $i=1,...,|G|$. The next step, as described above, is to estimate the one-dimensional probability distribution of such a projection, that is, the distribution of the outputs of the simple cells.  It is generally assumed that complex cells pool the outputs of simple cells.  Thus a complex cell could compute $\mu_n^k(I) =1/|G| \sum_{i=1}^{|G|} \sigma(\scal{I}{g_i t^k} + n\Delta)$ where $\sigma$ is a smooth version of the step function ($\sigma(x)=0$ for $x\le 0$, $\sigma(x)=1$ for $x > 0$) and $n=1,...,N$ (this corresponds to the choice $\eta_{n}(\cdot)\equiv \sigma(\cdot + n\Delta)$) . Each of these $N$ complex cells would estimate one bin of an approximated CDF (cumulative distribution function) for $P_{\scal{I}{t^k}}$. Following the theoretical arguments above, the complex cells could compute, instead of an empirical CDF, one or more of its moments. $n=1$ is the mean of the dot products, $n=2$ corresponds to an energy model of complex cells \cite{adelson1985spatiotemporal}; very large $n$ corresponds to a $max$ operation. Conventional wisdom interprets available physiological data to suggest that simple/complex cells in V1 may be described in terms of energy models, but our alternative suggestion of empirical histogramming by sigmoidal nonlinearities with different offsets may fit the diversity of data even better.

As described above, a template and its transformed versions may be learned from unsupervised visual experience through Hebbian plasticity. Remarkably, our analysis and empirical studies\cite{MM2013} show that Hebbian plasticity, as formalized by Oja, can yield {\it Gabor-like tuning}---i.e., the templates that provide optimal invariance to translation and scale (see SI Appendix section 2).

The localization condition (Equation \ref{locm}) can also be satisfied by images and templates that are similar to each other. The result is invariance to class-specific transformations. This part of the theory is consistent with the existence of class-specific modules in primate cortex such as a face module and a body module \cite{ kanwisher2010functional,Tsao2003,Leibo2011b}. It is intriguing that {\it the same localization condition} suggests {\it general Gabor-like templates for generic images} in the first layers of a hierarchical architecture and {\it specific, sharply tuned templates} for the last stages of the hierarchy. This theory also fits physiology data concerning Gabor-like tuning in V1 and possibly in V4 (see \cite{MM2013}). It can also be shown that the theory, together with the hypothesis that storage of the templates takes place via Hebbian synapses, also predicts properties of the tuning of neurons in the face patch AL of macaque visual cortex \cite{MM2013,Leibo2013}.

From the point of view of neuroscience, the theory makes a number of predictions, some obvious, some less so. One of the main predictions is that simple and complex cells should be found in all visual and auditory areas, not only in V1.  Our definition of simple cells and complex cells is different from the traditional ones used by physiologists; for example, we propose a broader interpretation of complex cells, which in the theory represent invariant measurements associated with histograms of the outputs of simple cells or of moments of it.  The theory implies that invariance to all image transformations could be learned, either during development or in adult life. It is, however, also consistent with the possibility that basic invariances may be genetically encoded by evolution but also refined and maintained by unsupervised visual experience. Studies on the development of visual invariance in organisms such as mice raised in virtual environments could test these predictions.

\section{Discussion}
The goal of this paper is to introduce a new theory of learning invariant representations for object recognition which cuts across levels of analysis \cite{MM2013, Marr1976}. At the computational level, it gives a unified account of \textit{why} a range of seemingly different models have recently achieved impressive results on recognition tasks. HMAX \cite{Riesenhuber1999a,  Mutch2006, Serre2007}, Convolutional Neural Networks \cite{Fukushima1980, LeCun1989, lecun1995convolutional, LeCun2004} and Deep Feedforward Neural Networks \cite{icassp12cnn, NIPS2012HintonImagenet, Le_highlevel2011} are examples of this class of architectures---as is, possibly, the feedforward organization of the ventral stream.  At the algorithmic level, it motivates the development, now underway, of a new class of models for vision and speech which includes the previous models as special cases. At the level of biological implementation, its characterization of the optimal tuning of neurons in the ventral stream is consistent with the available data on Gabor-like tuning in V1\cite{MM2013} and the more specific types of tuning in higher areas such as in face patches.

Despite significant advances in sensory neuroscience over the last five decades, a true understanding of the basic functions of the ventral stream in visual cortex has proven to be elusive. Thus it is interesting that the theory of this paper follows from a novel hypothesis about the main computational function of the ventral stream: the representation of new objects/images in terms of a signature which is invariant to transformations learned during visual experience, thereby allowing recognition from very few labeled examples---in the limit, just one.
A main contribution of our work to machine learning is a novel theoretical framework for the next major challenge in learning theory beyond the supervised learning setting which is now relatively mature: the problem of {\it representation learning}, formulated here as the unsupervised learning of invariant representations that significantly reduce the sample complexity of the supervised learning stage.

\begin{acknowledgments}
We would like to thank the McGovern Institute for Brain Research for their support. We would also like to thank for
having read earlier versions of the manuscript Yann LeCun, Ethan Meyers, Andrew Ng, Bernhard Schoelkopf and Alain Yuille.
We also thanks Michael Buice, Charles Cadieu, Robert Desimone, Leyla Isik, Christof Koch, Gabriel Kreiman,  Lakshminarayanan Mahadevan, Stephane Mallat, Pietro Perona, Ben Recht, Maximilian Riesenhuber, Ryan Rifkin,  Terrence J. Sejnowski, Thomas Serre, Steve Smale, Stefano Soatto, Haim Sompolinsky, Carlo Tomasi, Shimon Ullman and Lior Wolf for useful comments. This material is based upon work supported by the
Center for Brains, Minds and Machines (CBMM), funded by NSF STC award CCF-1231216.  This research was also sponsored by grants from the National Science Foundation (NSF-0640097, NSF-0827427), and AFSOR-THRL (FA8650-05-C-7262). Additional support was provided by the Eugene McDermott Foundation.
\end{acknowledgments}

\newpage

\section{Supplementary Information}

\section*{0.Invariance significantly reduces sample complexity}
In this section we show how, in the simple case  of  transformations which are translations,
an invariant representation of the image space considerably reduces the sample complexity of the classifier.\\
If we view images as vectors in $\R^d$, the sample complexity of a learning rule depends on the covering number of the ball, $B\in\R^d$, that
contains all the image distribution.   More precisely,  the covering number, $\mathcal{N}(\epsilon,B)$, is defined as the minimum
number of $\epsilon-$ balls  needed to cover $B$.  Suppose $B$ has radius $r$ we have
$$
\mathcal{N}(\epsilon,B)\sim\Big(\frac{r}{\epsilon}\Big)^{d}.
$$
For example, in the case of linear learning rules, the sample complexity is proportional to the {\em logarithm} of the covering number.

Consider the simplest and most intuitive example: an image made of a single pixel and its translations in a square of dimension $p\times p$, where $p^2=d$. In the pixel basis the space of the image and all its translates has dimension $p^2$ meanwhile the image dimension is one. The associated covering numbers are therefore
$$
\mathcal{N}^{I}(\epsilon,B)=\Big(\frac{r}{\epsilon}\Big)^{1},\;\;\;\mathcal{N}^{TI}(\epsilon,B)=\Big(\frac{r}{\epsilon}\Big)^{p^2}
$$
where  $\mathcal{N}^{I}$ stands for the covering number of the image space and $\mathcal{N}^{TI}$
 the covering number of the translated image space. The sample complexity associated to the image space (see e.g. \cite{CucSma02}) is  $O(1)$ and that associated to the translated images  $O(p^2)$. The sample complexity reduction of an invariant representation is therefore given by
$$
m_{inv}=O(p^2)=\frac{m_{image}}{p^2}.
$$
The above reasoning is independent on the choice of the basis since it depends only on   the dimensionality of the ball containing all the images.   For example we could have determined the dimensionality looking the cardinality of eigenvectors (with non null eigenvalue) associated to a circulant matrix of dimension $p\times p$ i.e. using the Fourier basis. In the simple case above, the cardinality is clearly $p^2$.\\
In general any transformation of an abelian group can be analyzed using the Fourier transform on the group. We conjecture that a similar reasoning holds for locally compact groups using a wavelet representation instead of the Fourier representation.\\
The example and ideas above leads to the following theorem:
\begin{theorem}
Consider a space of images of dimensions $p\times p$ pixels which may appear in any position within a window of size $rp\times rp$ pixels.
The usual image representation yields a sample complexity (of a linear classifier) of order $m_{image} = O(r^2p^2)$; the invariant representation yields (because of much smaller covering numbers) a  sample complexity of order
$$
m_{inv}= O(p^2)=\frac{m_{image}}{r^2}.
$$
\end{theorem}

\section*{1. Setup and Definitions}
Let $\XX$ be a Hilbert space with norm and inner product denoted by
$\nor{\cdot}$ and $\scal{\cdot}{\cdot}$, respectively.  We can think
of $\XX$ as the space of images (our images are usually ``neural
images''). We typically consider
$\XX=\reals^{d}$ or $L^{2}(\reals)$ or $L^{2}(\reals^{2})$.  We denote with $G$ a (locally) compact
group and with an abuse of notation, we denote by $g$ both a group element
in $G$ and its action/representation on $\XX$.\\
When useful we will make the following assumptions which are justified  from a biological point of view.

\noindent {\it Normalized dot products} of signals (e.g. images or ``neural
activities'') are usually assumed throughout the theory, for
convenience but also because they provide {\it the most elementary
  invariances -- to measurement units (origin and scale)}. We assume
that the dot products are between functions or vectors that are {\it
  zero-mean and of unit norm.} Thus $\scal{I}{t}$ sets $I =
\frac{I'- \bar{I'}}{\nor{I'- \bar{I'}}}$, $t = \frac{t' - \bar{t'}}{\nor{t' -
  \bar{t'}}}$ with $\bar{(\cdot)}$ the mean. This normalization stage before each dot product is
consistent with the convention that the empty surround
of an isolated image patch has zero value  (which can be taken to be the average
``grey'' value over the ensemble of images). In particular the dot product of a
template -- in general different from zero -- and the ``empty'' region
outside an isolated image patch will be zero. The dot product of two
uncorrelated images -- for instance of random 2D noise -- is also
approximately zero.\\
\noindent
{\bf Remarks:}
\begin{enumerate}
\item The $k$-th component of the signature associated with a
  {\it simple-complex} module is (see Equation \eqref{stabilityl} or \eqref{POG_disc_meas})
$\mu_{n}^{k}(I)=\frac{1}{|G_0|} \sum_{g\in { G}_{0}} \eta_{n}\Big(\scal{g I}{t^{k}}\Big)$
where the functions $\eta_n$ are such that $Ker(\eta_{n})=\{0\}$: in
  words, the empirical histogram estimated for $\scal{g
    I}{t^{k}}$ does not take into account the $0$ value, since it
  does not carry any information about the image patch.  The functions
  $\eta_n$ are also assumed to be positive and bijective.
\item Images $I$  have a
  maximum total possible support corresponding to a bounded region $B
  \subseteq \reals^{2}$, which we refer to as the \emph{visual field},
  and which corresponds to the spatial pooling range of the module at
  the top of the hierarchy of Figure $1$ in the main text.
  \emph{Neuronal images} are inputs to the modules
  in higher layers and are usually supported in a higher dimensional
  space, corresponding to the signature components provided by
  lower layers modules; \emph{isolated objects} are images with
  support contained in the pooling range of one of the modules at an
  intermediate level of the hierarchy. We use the notation $\nu
  (I),\mu (I)$ respectively for the simple responses
  $\scal{g I}{t^{k}}$ and for the complex response
  $\mu_{n}^{k}(I)=\frac{1}{|G_0|} \sum_{g\in { G}_{0}}
  \eta_{n}\Big(\scal{g I}{t^{k}}\Big)$. To simplify the notation we
  suppose that the center of the support of the signature at each layer $\ell$, $\mu_{\ell}(I)$, coincides
  with the center of the pooling range.
\item The domain of the dot products $\scal{g I}{t^{k}}$
  corresponding to templates and to  simple cells is in general
  different from the domain of the pooling  $\sum_{g\in { G}_{0}}
  $. We will continue to use the commonly used term {\it receptive
    field}  -- even if it mixes these two domains.

\item The main part of the theory characterizes properties of the
  basic HW module -- which computes the components of an
invariant signature vector from an image patch within
its receptive field.
\item It is important to emphasize that the {\it basic module is always the
  same} throughout the paper. We use different mathematical tools,
including approximations, to study under which conditions
(e.g. localization or linearization, see end of section 2) the signature computed
by the module is invariant or approximatively invariant.

\item The pooling $\sum_{g\in { G}_{0}},\;G_{0}\subseteq G$ is effectively over a {\it
    pooling window} in the group parameters. In the case of 1D scaling
  and 1D translations, the pooling window corresponds to an interval, e.g.
  $[a^j, a^{j+k}]$, of scales and an interval, e.g. $[-\bar{x}, \bar{x}]$, of $x$ translations,
  respectively.
\item All the results in this paper are valid in the case of a discrete or a
  continuous compact (locally compact) group: in the first case we have a sum over the
  transformations, in the second an integral over the Haar measure of the group.

\item Normalized dot products also eliminate the need of the explicit
  computation of the determinant of the Jacobian for affine
  transformations (which is a constant and is simplified dividing by the norms) assuring
  that $\scal{AI}{At}=\scal{I}{t}$, where $A$ is an affine transformation.

\end{enumerate}
\vspace{0.3in}

\section*{2. Invariance and uniqueness: Basic Module}
\subsection{Compact Groups (fully observable)}

Given an image $I\in\XX$ and a group representation $g$, the orbit
$O_I=\{I'\in\XX ~~s.t.~~I'=gI, g\in G \}$ is uniquely associated to an
image and all its transformations.  The orbit provides an invariant
representation of $I$, i.e. $O_I=O_{gI}$ for all $g\in G$.  Indeed, we
can view an orbit as all the possible realizations of a random variable
with distribution $P_I$ induced by the group action.  From this
observation, a signature $\Sigma(I)$  can be derived for compact groups,
by using results characterizing probability distributions via their
one dimensional projections.\\
\noindent
In  this section we study the signature  given by
$$
\Sigma(I)=(\mu^1(I),\dots,\mu^k(I))=(\mu_1^1(I),..,\mu_{N}^{1},..,\mu^{K}_{1},..,\mu^K_N(I) ),
$$
where each component $\mu^k(I)\in\reals^{N}$ is a histogram corresponding
to a one dimensional projection defined by a template $t^k\in\XX$.  In
the following we let $\XX=\reals^{d}$.

\subsection{Orbits and probability distributions}
 If $ G$ is a compact group,  the associated Haar measure $dg$ can be normalized to be a probability measure, so that, for any $I\in \reals^d$, we can define the random variable,
$$
Z_I:G\to \reals^d, \quad \quad Z_I(g)=gI.
$$
The corresponding distribution  $P_I$ is defined as $P_I(A)=dg(Z_I^{-1}(A))$ for any Borel set $A\subset \reals^d$ (with some abuse of notation we let $dg$ be the normalized Haar measure).\\
Recall that we define two images, $I,I'\in\XX$ to be equivalent (and
we indicate it with $I\sim I'$) if there exists $g\in G$
s.t. $I=gI'$. We have the following theorem:\\

\begin{theorem}\label{PD}
The distribution $P_{I}$ is invariant and unique i.e. $I\sim I' \;\Leftrightarrow\; P_{I}=P_{I'}$.
\end{theorem}
\noindent
{\bf Proof:}\\
We first prove that $I\sim I' \;\Rightarrow\; P_{I}=P_{I'}$.
By definition $P_{I}=P_{I'}$ iff  $
\int_{A}dP_{I}(s)=\int_{A}dP_{I'}(s)$, $\forall\;A\subseteq {\cal X}$,   that is  $\int_{Z^{-1}_{I}(A)}dg = \int_{Z^{-1}_{I'}(A)}dg$, where,
\begin{eqnarray*}
&& Z^{-1}_{I}(A)=\{{g\in G}\;s.t.\;gI\subseteq A\}\\
&& Z^{-1}_{I'}(A)=\{{g\in  G}\;s.t.\;gI'\in A\}=\{g\in { G}\;s.t.\;g\bar{g}I\subseteq A\},
\end{eqnarray*}
$\forall\;A\subseteq {\cal X}$. Note that $\forall\;A\subseteq {\cal X}$ if $gI\in A\;\Rightarrow\; g\bar{g}^{-1}\bar{g}I= g\bar{g}^{-1}I'\in A$, so that $g\in Z^{-1}_{I}(A)\;\Rightarrow\;g\bar{g}^{-1}\in Z^{-1}_{I'}(A)$, i.e. $Z^{-1}_{I}(A)\subseteq Z^{-1}_{I'}(A)$. Conversely  $g\in Z^{-1}_{I'}(A)\;\Rightarrow\;g\bar{g}\in Z^{-1}_{I}(A)$, so that $Z^{-1}_{I}(A)=Z^{-1}_{I'}(A)\bar{g},\;\forall A$.  Using this observation we have,
$$
\int_{Z^{-1}_{I}(A)}dg = \int_{(Z^{-1}_{I'}(A))\bar{g}}dg =  \int_{Z^{-1}_{I'}(A)}d\hat{g}
$$
where in the last integral we used the change of variable $\hat{g}=g\bar{g}^{-1}$ and the invariance property of the Haar measure: this proves the implication.\\
To prove that $P_{I}=P_{I'}\;\Rightarrow\; I\sim I'$, note that $P_{I}(A)-$ $P_{I'}(A)=0$ for some $A\subseteq {\cal X}$ implies that the support of the probability distribution of $I$ has non null intersection with that of $I'$ i.e. the orbits of $I$ and $I'$ intersect.
In other words  there exist $g',g''\in {G}$ such that $g'I=g''I'$. This implies $I={g'}^{-1}g''I'=\bar{g}I',\;\bar{g}={g'}^{-1}g''$, i.e. $I\sim I'$. Q.E.D.\\

\subsection{Random Projections for Probability Distributions.}\label{CWsm}

Given the above discussion, a \textit{signature} may be associated to $I$ by  constructing a histogram approximation of  $P_I$,
but this would require dealing with high dimensional histograms. The following classic theorem gives a way around this problem.\\
For a {\em template} $t\in \mathbb{S}(\reals^{d})$,
where $\mathbb{S}(\reals^{d})$ is unit sphere  in $\reals^{d}$, let $I\mapsto \scal{I}{t}$ be   the associated projection.
Moreover,  let $P_{\scal{I}{t}}$ be the distribution associated to the
random variable $g\mapsto \scal{gI}{t}$ (or equivalently $g\mapsto \scal{I}{g^{-1}t} $, if $g$ is unitary).
Let $\mathcal{E}=[t\in \mathbb{S}(\reals^{d}),\;s.t.\;P_{\scal{I}{t}  }= Q_{\scal{I}{t} }  ]$. \\
\begin{theorem} (Cramer-Wold, \cite{CramerWold1936})
For any  pair $P,Q$ of probability distributions on $\reals^{d}$, we have that
$P =Q$ if and only if $\mathcal{E}=\mathbb{S}(\reals^{d})$.
\end{theorem}
In words, two probability distributions are equal if and only if their projections on any of the unit sphere directions is equal.
The above result can be equivalently stated as saying that the probability of choosing
$t$ such that $P_{\scal{I}{t}} = Q_{\scal{I}{t}}$ is equal to 1 if and only if $P = Q$ and  the
probability of choosing $t$ such that $P_{\scal{I}{t}} = Q_{\scal{I}{t}}$ is equal to 0 if
and only if $P \ne Q$ (see Theorem 3.4 in \cite{Cuesta2007}).
The theorem suggests a way to define a metric on distributions (orbits) in
terms of
$$
d(P_I,P_{I'})=\int d_0(P_{\scal{I}{t}},P_{\scal{I'}{t}}) d\lambda(t), \quad \forall I,I' \in {\cal X},
$$
where $d_0$ is any metric on one dimensional probability
distributions and $d\lambda(t)$ is a distribution
measure on the projections.  Indeed, it is easy to check that $d$ is a metric. In
particular note that, in view of the Cramer Wold Theorem, $d(P,Q)=0$ if
and only if $P=Q$.  As mentioned in the main text, each one dimensional distribution
$P_{\scal{I}{t}}$ can be approximated by a suitable histogram
$\mu^t(I)=(\mu_n^t(I))_{n=1,\dots,N}\in R^N$, so that, in the limit in
which the histogram approximation is accurate
\begin{equation}\label{hsliced}
d(P_I,P_{I'}) \approx \int d_\mu(\mu^t(I),\mu^t(I')) d\lambda(t), \quad \forall I,I' \in {\cal X},
\end{equation}
where $d_\mu$ is a metric on histograms induced by $d_0$.

A natural question is whether there are situations in which a finite
number of projections suffice to discriminate any two probability
distributions, that is $P_I\neq P_I'\Leftrightarrow d(P_I, P_{I'})\neq
0$.  Empirical results show that this is often the case with a small
number of templates (see \cite{Cuesta2009} and HMAX experiments, section 6).  The
problem of mathematically characterizing the situations in which a finite
number of (one-dimensional) projections are sufficient is challenging.
Here we  provide a partial answer to this question.\\
We start by observing that the metric~\eqref{hsliced} can be approximated
by uniformly sampling $K$ templates and considering
\begin{equation}\label{sliced}
\hat{d}_K(P_I,P_{I'})=\frac 1 K  \sum_{k=1}^K d_\mu(\mu^k(I),\mu^k(I')),
\end{equation}
where $\mu^k=\mu^{t^k}$. The following result shows that a finite
number $K$ of templates is sufficient to obtain an approximation
within a given precision $\epsilon$. Towards this end let
\begin{equation}\label{empsliced}
d_\mu(\mu^k(I),\mu^k(I'))=\nor{\mu^k(I)-\mu^k(I')}_{\reals^N}.
\end{equation}
where $\nor{\cdot}_{\reals^{N}}$ is the Euclidean norm in
$\reals^{N}$. The following theorem holds:\\
\begin{theorem}
\label{akaJL}
Consider $n$ images ${\cal X}_n$ in $\cal X$. Let $K \ge
\frac{2}{c\epsilon^2} \log {\frac{n}{\delta}}$, where $c$ is a
universal constant.  Then
\begin{equation} \label{CWJL}
|d(P_I,P_{I'})
-\hat{d}_K(P_I,P_{I'})|\le\epsilon,
\end{equation}
with probability $1 - \delta^2$, for all  $I,I'\in {\cal X}_n$.
\end{theorem}
\noindent
{\bf Proof:}\\
The proof follows from an application of H\"oeffding's inequality and a union bound.\\
Fix $I,I'\in \XX_{n}$. Define the real random
variable $Z:\mathbb{S}(\R^{d}) \to \R$,
$$
Z(t^k)=\nor{\mu^k(I)-\mu^k(I')}_{\R^N},\;\;k=1, \dots, K.
$$
From the
definitions it follows that
$\nor{Z}\le c$  and $\mathbb E (Z)=d(P_I, P_{I'})$. Then H\"oeffding
inequality implies
$$ |d(P_I, P_{I'}) -\hat{d}_K(P_I, P_{I'})|= |\frac 1 K \sum_{k=1}^{K} \mathbb E (Z) - Z(t^{k}) | \ge \epsilon,$$
with probability at most $e^{-c\epsilon^2k}$.  A union bound implies a
result holding uniformly on $\XX_{n}$; the
probability becomes at most $n^2e^{-c\epsilon^2K}$.  The desired
result is obtained noting that this probability is less than $\delta^2$
as soon as $n^2e^{-c\epsilon^2K}<\delta^2$ that is $K \ge
\frac{2}{c\epsilon^2} \log {\frac{n}{\delta}}$. Q.E.D.\\\\
\noindent
The above result shows that the discriminability question can be
answered in terms of empirical estimates of the one-dimensional
distributions of projections of the image and transformations induced
by the group on a number of templates $t^k, ~ k=1,...,K$.\\
Theorem 4 can be compared to a version of the Cramer Wold Theorem for
discrete probability distributions. Theorem 1 in \cite{Heppes1956}
shows that for a probability distribution consisting of $k$ atoms in
$\reals^{d}$, we see that at most $k + 1$ directions
($d_{1}=d_{2}=...=d_{k+1}=1$) are enough to characterize the
distribution, thus a finite -- albeit large -- number of
one-dimensional projections.

\subsection{Memory based learning of invariance}
The signature $\Sigma(I)=(\mu_1^1(I), \dots, \mu^K_N(I) )$ is
  obviously invariant (and unique) since it is associated to an image
  and all its transformations (an orbit).  Each component of the
  signature is also invariant -- it corresponds to a group
  average. Indeed, each measurement can be defined as
\begin{equation}\label{disc_meas}
 \mu_{n}^{k}(I)=\frac{1}{|G|} \sum_{g\in { G}} \eta_{n}\Big(\scal{g I}{t^{k}}\Big),
\end{equation}
for $G$  finite group, or equivalently
\begin{equation}\label{cont_meas}
 \mu_{n}^{k}(I)=\int_{G} dg\;  \eta_{n}\Big(\scal{g I}{t^{k}}\Big)=\int_{G} dg \; \eta_{n}\Big(\scal{ I}{g^{-1}t^{k}}\Big),
\end{equation}
when $G$ is a (locally) compact group. Here, the non linearity $\eta_n$ can be chosen to define an histogram approximation; in general is a bijective positive function.
Then, it is clear that from the properties of the Haar measure we have
\begin{equation}\label{inv_lem}
\mu_{n}^{k}(\bar{g}I)= \mu_{n}^{k}(I), \quad\forall \bar{g}\in G, I\in \XX.
\end{equation}
Note that in the r.h.s. of eq. \eqref{cont_meas} the transformations are on templates: this mathematically trivial (for unitary transformations) step has a deeper computational aspect. Invariance is now in fact achieved  through transformations of templates instead of those of the image, not always available.

\subsection{Stability}

With $\Sigma(I)\in \reals^{NK}$ denoting as usual the signature of an
image, and $d(\Sigma(I),\Sigma(I'))$, $I,I'\in \XX$, a metric, we
say that a {\it signature $\Sigma$ is stable if it is Lipschitz
  continuous} (see \cite{MallatGroupInvariantMain}), that is
\begin{equation}\label{stabilityl}
d(\Sigma(I),\Sigma(I')) \le L \nor{I-I'}_{2}, \quad L>0,\;\;\forall I,I'\in {\cal X}.
\end{equation}
In our setting we let
$$d(\Sigma(I),\Sigma(I'))=\frac 1 K \sum_{k=1}^{K}d_\mu(\mu^k(I),\mu^k(I')),$$
and assume that $\mu_{n}^{k}(I)=\int dg \;\eta_{n}(\scal{gI}{t^{k}})$
for $n=1, \dots, N$ and $k=1,\dots, K$. If $L<1$ we call the signature map contractive. In the following we prove a stronger form of eq. \ref{stabilityl} where the $L^{2}$ norm is substituted with the Hausdorff norm on the orbits (which is independent of the choice of $I$ and $I'$ in the orbits) defined as $\nor{I-I'}_H= min_{g,g'\in G}\nor{gI-g'I'}_2$,  $I,I'\in \XX$, i.e. we have:

\begin{theorem}\label{St} Assume normalized templates and let $L_{\eta}= \max_{n}
  (L_{\eta_{n}})$ s.t. $NL_{\eta}\leq 1$, where $L_{\eta_{n}}$ is the Lipschitz constant
  of the function $\eta_{n}$. Then
\begin{equation}\label{Stab}
d(\Sigma(I),\Sigma(I')) < \nor{I-I'}_H,
\end{equation}
 for all $I,I'\in \XX$.
\end{theorem}
\noindent
{\bf Proof:}\\
By definition, if the  non linearities $\eta_{n}$ are  Lipschitz continuous, for all $n=1, \dots, N$, with
Lipschitz constant $L_{\eta_{n}}$, it follows that for each $k$ component of the signature we have
\begin{eqnarray*}
&& \nor{\Sigma^k(I)-\Sigma^k(I')}_{\R^{N}}\\
&\leq& \frac{1}{|G|}\sqrt{\sum_{n=1}^{N}\Big(\sum_{g\in G}L_{\eta_{n}}|\scal{gI}{t^{k}}-\scal{gI'}{t^{k}}|\Big)^{2}}\\
&\leq& \frac{1}{|G|}\sqrt{\sum_{n=1}^{N}L^{2}_{\eta_{n}}\sum_{g\in G}(|\scal{g(I-I')}{t^{k}}|)^{2}},
\end{eqnarray*}
where we used the linearity of the inner product and Jensen's inequality. Applying Schwartz's inequality we obtain
$$
\nor{\Sigma^k(I)-\Sigma^k(I')}_{\R^{N}} \leq \frac{L_{\eta}}{|G|}\sqrt{\sum_{n=1}^{N}\sum_{g\in G}\nor{I-I'}^{2}\nor{g^{-1}t^{k}}^{2}}
$$
where $L_{\eta}= \max_{n} (L_{\eta_{n}})$. If we assume the  templates and their transformations to be
normalized to unity then we finally have,
\begin{equation}\label{stability2}
\nor{\Sigma^k(I)-\Sigma^k(I')}_{\R^{N}}\leq N L_{\eta} \nor{I-I'}_{2}.
\end{equation}
from which we obtain  \eqref{stabilityl} summing over all $K$ components and dividing by $1/K$ since $NL_{\eta}<1$ by hypothesis. Note now that the l.h.s. of \eqref{stability2}, being each component of the signature $\Sigma(\cdot)$ invariant, is independent of the choice of $I,I'$ in the orbits. We can then choose $\tilde{I},\tilde{I}'$ such that
$$
\nor{\tilde{I}-\tilde{I}'}_{2} =  min_{g,g'\in G}\nor{gI-g'I'}_2=\nor{I-I'}_H
$$
In particular being $NL_{\eta}< 1$
the map is non expansive summing each component and dividing by $1/K$ we have eq. \eqref{Stab}. Q.E.D.\\\\
The above result shows
that the stability of the empirical signature
$$\Sigma(I)=(\mu_1^1(I), \dots, \mu^K_N(I) )\in \reals^{NK},$$
provided with the metric~\eqref{sliced} (together
with~\eqref{empsliced}) holds for nonlinearities with Lipschitz
constants $L_{\eta_{n}}$ such that $N max_{n}(L_{\eta_{n}})<1$.

\vspace{0.5 in}

\begin{framed}
\captionof*{algorithm}{\it Box 1:  computing an invariant signature $\mu(I)$}
\begin{algorithmic}[1]
\Procedure{\texttt{Signature}} {I}  \label{LearningInvariance}
    \Statex Given $K$ templates $\{ g t^k | \forall g \in G\}$.
    \For{$k = 1, \dots, K$}
    \State \begin{varwidth}[t]{\linewidth} Compute $\scal{I}{g t^k}$, the normalized \newline dot products of the image with all the \newline transformed  templates (all $g \in G$).
    \end{varwidth}
    \State Pool the results: \texttt{POOL}($\{\scal{I}{g t^k}|\forall g \in G\}$).
    \EndFor
    \State \textbf{return} $\mu(I) =$ the pooled results for all $k$. \hspace{0.2in} \Comment{$\mu(I)$ is unique and invariant if there are enough templates.}
\EndProcedure \newline
\end{algorithmic}
\end{framed}

\subsection{Partially Observable Groups case: invariance implies localization and sparsity}

This section outlines invariance, uniqueness and stability properties
of the signature obtained in the case in which transformations of a
group are observable only within a {\it window} ``over'' the
orbit. The term POG (Partially Observable Groups) emphasizes the
properties of the group -- in particular associated invariants -- as
seen by an observer (e.g. a neuron) looking through a window at a part
of the orbit.
Let $G$ be a finite group and
$G_0\subseteq G$ a subset (note: $G_0$ is not usually a subgroup). The subset of transformations $G_0$ can be
seen as the set of transformations that can be {\em observed} by a
window on the orbit that is the transformations that correspond to a
part of the orbit.  A {\em local} signature associated to the partial
observation of $G$ can be defined considering
\begin{equation}\label{POG_disc_meas}
 \mu_{n}^{k}(I)=\frac{1}{|G_0|} \sum_{g\in { G}_{0}} \eta_{n}\Big(\scal{g I}{t^{k}}\Big),
\end{equation}
and $\Sigma_{G_0}(I)=( \mu_{n}^{k}(I))_{n,k}$.
This definition can be generalized to any locally compact group considering,
\begin{equation}\label{mu}
\mu_{n}^{k}(I)=\frac{1}{V_{0}} \int_{{ G}_{0}} \eta_{n}\Big(\scal{g I}{t^{k}}\Big)dg, \quad  V_0=\int_{G_0} dg.
\end{equation}
Note that the constant $V_0$ normalizes the Haar measure, restricted
to $G_0$, so that it defines a probability distribution.  The latter is the
distribution of the images subject to the group transformations which
are observable, that is in $G_0$.  The above definitions can be
compared to definitions \eqref{disc_meas} and \eqref{cont_meas} in the
fully observable groups case.  In the next sections we discuss the
properties of the above signature. While stability and uniqueness
follow essentially from the analysis of the previous section,
invariance requires developing a new analysis.

\subsection{POG: Stability and Uniqueness} A direct consequence of
Theorem~\ref{PD} is that {\it any two orbits with a common point are
  identical}. This follows from the fact that if  $gI,g'I'$ is a common point of the orbits, then
$$
g'I'=gI\;\Rightarrow\;I'= (g')^{-1}gI.
$$
Thus the two images are transformed versions of one another and $O_{I}=O_{I'}$.\\
Suppose now that only a fragment of the orbits -- the part within the
window -- is observable; the reasoning above is still valid since if
the orbits are different or equal so must be any of their
``corresponding'' parts. \\
Regarding the stability of POG signatures, note that the reasoning in the
previous section, Theorem \ref{St}, can be repeated without any significant change. In
fact, only the normalization over the transformations is modified
accordingly.


\subsection{POG: Partial Invariance and Localization}
\begin{figure}
\centering
\includegraphics[width= 9cm]{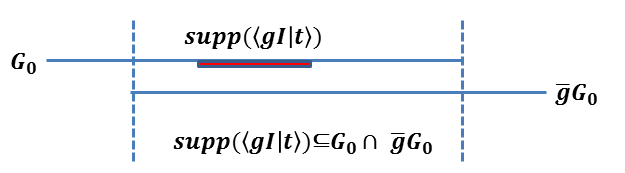}
\caption{A sufficient condition for invariance for locally compact groups: if the support of $\scal{gI}{t}$ is sufficiently localized it will be completely contained in the pooling interval even if the image is group shifted, or, equivalently (as shown in the Figure), if the pooling interval is group shifted by the same amount.\label{LCG_condition}}
\end{figure}

Since the
group is only partially observable we introduce the notion of {\em partial
  invariance} for images and transformations $G_0$ that are within the
observation window. Partial invariance is defined in terms of
invariance of
\begin{equation}
 \mu_{n}^{k}(I)=\frac{1}{V_{0}} \int_{ G_0}\;dg\; \eta_{n}\Big(\scal{g I}{t^{k}}\Big).
\end{equation}
We recall that when $g I$ and $t^{k}$ do not
share any common support on the plane or $I$ and $t$ are uncorrelated, then
$\scal{g I}{t^{k}} = 0$.
The following theorem, where $G_{0}$ corresponds to the pooling
range states, a sufficient  condition for partial invariance in the case of a locally compact group:\\
\begin{theorem}{\bf Localization and Invariance.}\label{generalinv}
Let $I,t\in H$ a Hilbert space, $\eta_{n}:\R\rightarrow\R^{+}$ a set of bijective (positive) functions and $G$ a locally compact group. Let $G_{0}\subseteq G$ and suppose $supp(\scal{gI}{t^k})\subseteq G_{0}$. Then for any given $\bar{g}\in G$,  $t^k,I\in \XX$ the following conditions hold:
\begin{eqnarray}\label{loc}
&&\scal{gI}{t^{k}}=0,\; \forall g\in G/(G_{0}\cap\bar{g}G_{0})\nonumber\\
&&\textrm{or equivalently}\;\;\;\;\;\;\;\;\;\;\;\;\;\;\;\;\;\;\;\;\;\;\;\;\;\Rightarrow\;\mu^{k}_{n}(I) = \mu^{k}_{n}(\bar{g}I)\nonumber \\
&&\scal{gI}{t^{k}}\neq 0,\; \forall g\in G_{0}\cap\bar{g}G_{0}
\end{eqnarray}
\end{theorem}
{\bf Proof:}\\
\noindent
To prove the implication note that if $\scal{gI}{t^{k}}=0,\;\forall g\in G/(G_{0}\cap\bar{g}G_{0})$, being $G_{0}\Delta \bar{g}G_{0}\subseteq G/(G_{0}\cap\bar{g}G_{0})$ ($\Delta$ is the symbol for symmetric difference ($ A\Delta B = (A\cup B)/ (A\cap B)\;\;A,B\;\;sets$) we have:
\begin{eqnarray}\label{loc_inv}
0 &=& \int_{G/(G_{0}\cap\bar{g}G_{0})}dg\;\eta_{n} \big(\scal{gI}{t^{k}}\big)\nonumber \\
  &=& \int_{G_{0}\Delta\bar{g}G_{0}}dg\;\eta_{n} \big(\scal{gI}{t^{k}}\big)\\
  &\geq& |\int_{G_{0}}dg\;\Big(\eta_{n}\big(\scal{gI}{t^{k}}\big)-\eta_{n}\big(\scal{g\bar{g}I}{t^{k}}\big)\Big)|. \nonumber
\end{eqnarray}
 The second equality is true since, being $\eta_{n}$ positive, the fact that the integral is zero implies $\scal{gI}{t^{k}}=0\;\forall g\in G/(G_{0}\cap\bar{g}G_{0})$ (and therefore in particular $\forall g\in G_{0}\Delta\bar{g}G_{0}$). Being the r.h.s. of the inequality positive, we have
\begin{equation}
|\int_{G_{0}}dg\;\Big(\eta_{n}\big(\scal{gI}{t^{k}}\big)-\eta_{n}\big(\scal{g\bar{g}I}{t^{k}}\big)\Big)|=0
\end{equation}
i.e. $\mu^{k}_{n}(I) = \mu^{k}_{n}(\bar{g}I)$ (see also Fig. \ref{LCG_condition}  for a visual explanation). Q.E.D.\\
\noindent
Equation  \eqref{loc} describes a \emph{localization} condition on the inner product of the transformed image and the template.
The above result naturally raises question of weather the localization condition is also necessary for invariance.
Clearly, this would  be the case if eq. \eqref{loc_inv} could be turned into an equality,  that is
\begin{eqnarray}\label{condelta}
&&\int_{G_{0}\Delta\bar{g}G_{0}}dg\;\eta_{n} \big(\scal{gI}{t^{k}}\big)\\
&=& |\int_{G_{0}}dg\;\Big(\eta_{n}\big(\scal{gI}{t^{k}}\big)-\eta_{n}\big(\scal{g\bar{g}I}{t^{k}}\big)\Big)|\nonumber\\
&=& |\mu^{k}_{n}(I) - \mu^{k}_{n}(\bar{g}I)|\nonumber.
\end{eqnarray}
Indeed,  in this case, if $\mu^{k}_{n}(I) - \mu^{k}_{n}(\bar{g}I)=0$, and we further assume  the natural condition $\scal{gI}{t^k} \neq 0$ if and only if $g\in G_0$, then the localization condition  \eqref{loc} would be necessary since $\eta_{n}$ is a positive bijective function.\\
The equality in eq. \eqref{condelta}  in general is not  true. However, this is clearly the case if we consider the group of transformations to be translations as illustrated in  Fig. \ref{FigInvdt} a). We discuss in some details  this latter case.

\noindent
Assume that $G_0=[0,a]$. Let
\begin{equation}
S=\{\scal{T_{x}I}{t}~:~\;x\in[0,a]\},\;\; S_c= \{\scal{T_{x}I}{t}~:~\;x\in[c,a+c]\},
\end{equation}
for a given $c$ where $T_{x}$ is a unitary representation of the translation operator. We can view  $S,S_c$ as  sets of simple responses to a given template
through two  receptive fields.  Let $S_0=\{\scal{T_{x}I}{t}~:~\;x\in[0,a+c]\}$, so that $S,S_c\subset S_0$ for all $c$. We assume that $S_0,S,S_c$ to be closed intervals for all $c$. Then, recall that a bijective function (in this case $\eta_{n}$) is  strictly monotonic on any closed interval so that   the  difference of integrals in eq. \eqref{condelta} is zero if and only if $S=S_c$.  Since we are interested in considering all the values of  $c$ up to some maximum $C$, then we can  consider the  condition
\begin{equation}
\scal{T_{x}I}{t} = \scal{T_{x}T_{a}I}{t},\; \forall\;x\in[0,c],\;c\in[0,C].
\end{equation}
The above condition  can be  satisfied in two cases: 1) both dot products are zero, which is the localization condition, or 2) $T_{a}I=I$ (or equivalently $T_{a}t=t$) i.e. the image or the template are periodic. A similar reasoning applies to the case of scale transformations.

In the next paragraph we will see how localization conditions for scale and translation transformations imply a specific form of the templates.

\begin{figure} \begin{center} \includegraphics[scale=.33]{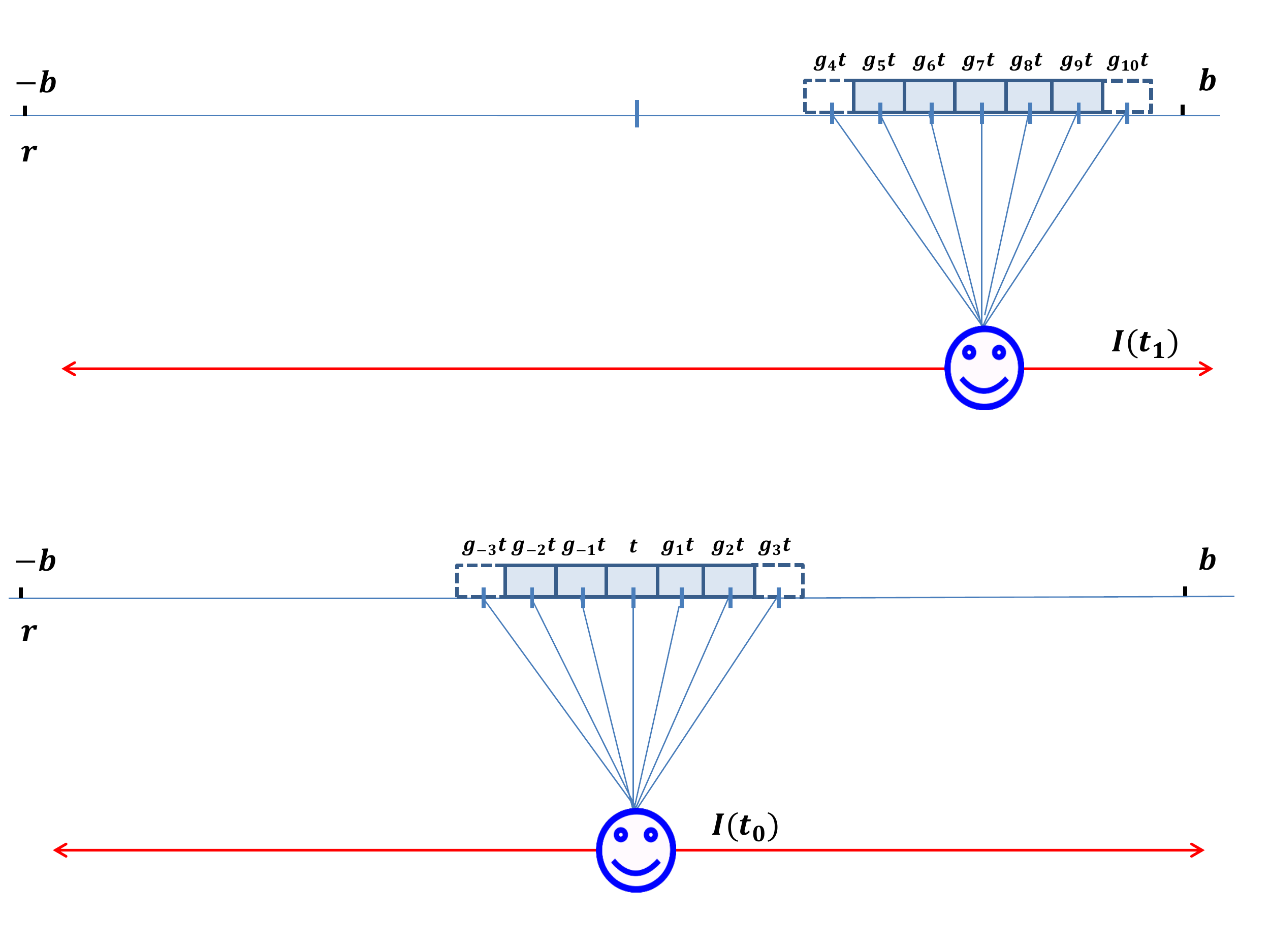} \end{center} \caption{An HW-module pooling the dot products of transformed templates with the image. The input image $I$  is shown centered on the template $t$; the same module is shown above for a group shift of the input image, which now localizes around the transformed template $g_{7}t$. Images and templates satisfy the localization condition $\scal{I}{T_{x}t}\neq 0,\;|x|>a$ with $a=3$. The interval $[-b,b]$ indicates the pooling window. The shift in $x$ shown in the Figure is a special case: the reader should consider the case in which the transformation parameter, instead of $x$, is for instance rotation in depth.
\label{RF_condition}}
\end{figure}

\vspace{0.1in}


\paragraph{The Localization condition: Translation and Scale}

In this section we identify $G_0$ with subsets
of the affine group. In particular, we study separately the case of
scale and translations (in 1D for simplicity).

In the following it is helpful to assume that all images $I$ and
templates $t$ are strictly contained in the range of translation or
scale pooling, $P$, since image components outside it are not measured. We
will consider images {\it $I$ restricted to $P$}: for translation this means that
the support of $I$ is contained in $P$, for
scaling, since $g_s I= I(sx)$ and $\widehat{I(sx)}=(1/s)\hat{I}(\omega/s)$ (where $\hat{\cdot}$ indicates the Fourier transform),
assuming a scale pooling range of $[s_m, s_M]$, implies a range
$[\omega^I_m, \omega^I_M],\;[\omega_m^t,\omega_M^t]$ ($m$ and $M$
indicates maximum and minimum) of spatial frequencies for the maximum support of
$I$ and $t$. As we will see because of Theorem \ref{generalinv} {\it invariance to
  translation requires spatial localization of images and templates} and
less obviously {\it invariance to scale requires bandpass properties
  of images and templates}. Thus images and templates are assumed to
be localized from the outset in either space or frequency. The
corollaries below show that a stricter localization condition is
needed for invariance and that this condition determines the form of
the template. Notice that in our framework images and templates are bandpass because of
  being zero-mean.  Notice that, in addition, neural ``images'' which
  are input to the hierarchical architecture are spatially bandpass
  because of retinal processing.

\noindent
We now state the result of Theorem \ref{generalinv} for one
dimensional signals under the translation group and -- separately -- under
the dilation group.\\
\hspace{0.1in}
Let $I,t \in L^{2}(\R)$, $(\R,+)$ the one dimensional locally
compact group of translations ($T_{x}:L^{2}(\R)\rightarrow
L^{2}(\R)$ is a unitary representation of the translation
operator as before). Let, e.g., $G_{0}=[-b,b],\;b>0$  and suppose $supp(t)\subseteq
supp(I)\subseteq [-b,b]$. Further suppose
$supp(\scal{T_{x}I}{t})\subseteq [-b,b]$. Then
eq. \eqref{loc} (and the following discussion for the translation (scale) transformations) leads to\\
\noindent {\bf Corollary 1:} {\it Localization in the spatial domain
  is necessary and sufficient for translation invariance.} For any
fixed $t,I\in\XX$ we have:
\begin{equation}\label{loct}
\mu^{k}_{n}(I)=\mu^{k}_{n}(T_{x}I),\;\forall x\in [0,\bar{x}]\;\Leftrightarrow\; \scal{T_{x}I}{t}\neq 0,\;\forall x\in[-b+\bar{x},b].
\end{equation}
with $\bar{x}>0$.\\
\noindent
Similarly let $G=(\R^{+},\cdot)$ be the one dimensional
locally compact group of dilations and denote with $D_{s}:L^{2}(\R)\rightarrow
L^{2}(\R)$ a unitary representation of the dilation operator. Let
$G_{0}=[1/S,S],\; S>1$ and suppose
$supp(\scal{D_{s}I}{t})\subseteq [1/S,S]$. Then\\
\noindent {\bf Corollary 2:} {\it Localization in the spatial frequency domain is necessary and sufficient for
  scale invariance.} For any fixed $t,I\in\XX$ we have:
\begin{equation}\label{locd}
\mu^{k}_{n}(I)=\mu^{k}_{n}(D_{s}I),s\in [1,\bar{s}]\;\;\Leftrightarrow\; \scal{D_{s}I}{t}\neq 0,\;\forall s\in [\frac{\bar{s}}{S},S].
\end{equation}
with $S>1$.\\
\noindent
Localization conditions of the support of the dot product for translation and scale are depicted in Figure \ref{FigInvdt},a) ,b).\\
As shown by the following Lemma \ref{supports} Eq. \eqref{loct} and
\eqref{locd} gives  interesting conditions on the supports
of $t$ and its Fourier transform $\hat{t}$.
For translation, the corollary is equivalent to zero overlap of the compact supports of
$I$ and $t$. In particular using Theorem \ref{generalinv}, for $I=t$, the maximal invariance in translation implies the following localization conditions on $t$
\begin{equation}\label{SD}
\scal{T_{x}t}{t} = 0 \;\; |x|>a,a>0
\end{equation}
which we call  self-localization.\\
For scaling we consider the support of the Fourier transforms of $I$
and $t$.
The Parseval's theorem allows to rewrite
the dot product $\scal{D_{s}I}{t}$ which is in $L^{2}(\reals^{2})$ as
$\scal{\widehat{D_{s}I}}{\hat{t}}$ in the Fourier domain.\\
In the following we suppose that the support
of $\hat{t}$ and $\hat{I}$ is respectively $[\omega^{t}_m, \omega^{t}_M]$ and $[\omega_{m}^{I},\omega_{M}^{I}]$ where $\omega^{t,I}_m$ could be very
close to zero (images and templates are supposed to be zero-mean) but usually are
bigger then zero.\\
Note that the effect of scaling $I$ with (typically $s=2^j$ with
$j \leq 0$) is to change the support as $supp(\widehat{D_{s}I})=s (supp(\hat{I}))$.\\
This change of the support of $\hat{I}$ in the dot product $\scal{\widehat{D_{s}I}}{\hat{t}}$ gives non trivial conditions on the intersection with the support of $\hat{t}$ and therefore on the localization w.r.t. the scale invariance. We have the following Lemma:
\vspace{0.1in}
\begin{lemma}\label{supports}
Invariance to translation in the range $[0,\bar{x}], \;\bar{x}>0$ is equivalent to the following localization condition of $t$ in space
\begin{equation}\label{loctsupp}
supp(t)\subseteq [-b-\bar{x},b]-supp(I),\;I\in \XX.
\end{equation}
Separately, invariance to dilations in the range $[1,\bar{s}],\;\bar{s}>1$ is equivalent to the following localization condition of
$\hat{t}$ in frequency $\omega$
\begin{eqnarray}\label{supphatt}
&& supp(\hat{t})\subseteq [-\omega_{t} -\Delta^*_{t},-\omega_{t}+\Delta^*_{t}]\cup [\omega_{t} -\Delta^*_{t},\omega_{t}+\Delta^*_{t}]\nonumber\\
&& \Delta^{*}_{t}=S\omega^{I}_{m}-\omega_{M}^{I}\frac{\bar{s}}{S},\;\;\omega_{t}=\frac{\omega^{t}_{M}-\omega^{t}_{m}}{2}.
\end{eqnarray}
\end{lemma}
\noindent {\bf Proof:}\\
\noindent To prove that $supp(t)\subseteq [-b+\bar{x},b]-supp(I)$ note that eq. \eqref{loct} implies that $supp(\scal{T_{x}I}{t})\subseteq [-b+\bar{x},b]$ (see Figure \ref{FigInvdt}, a)).
In general $supp(\scal{T_{x}I}{t})=supp(I*t)\subseteq supp(I)+supp(t)$. The inclusion account for the fact that the integral $\scal{T_{x}I}{t}$ can be zero even if the supports of $T_{x}I$ and $t$ are not disjoint. However
if we suppose invariance for a continuous set of translations $\bar{x}\in[0, \bar{X}]$, (where, for any given $I,t$, $\bar{X}$ is the maximum translation for which we have an invariant signature) and for a generic image in $\XX$ the inclusion become an equality, since for the invariance condition in Theorem
\ref{generalinv} we have
\begin{eqnarray*}
&& \scal{T_{\bar{x}}I}{T_{x}t} = \scal{I}{T_{x}T_{\bar{x}}t}=\int_{-\infty}^{+\infty} I(\xi)(T_{x}t(\xi+\bar{x}))d\xi=0\\
&&\forall x \in [-\infty,-b]\cup [b,\infty],\; \forall \bar{x} \in [0,\bar{X}],\;\forall I \in \XX
\end{eqnarray*}
which is possible, given the arbitrariness of $\bar{x}$  and $I$ only if
\begin{eqnarray*}
&& supp(I)\cap T_{x}T_{-\bar{x}}supp(t)=\emptyset\\
&& \forall \bar{x} \in [0,\bar{X}],\;\;\forall x \in [-\infty,-b]\cup [b,\infty]
\end{eqnarray*}
where we used the property $supp(T_{x}f)=T_{x}f,\forall f\in \XX$.
Being, under these conditions, $supp(\scal{T_{x}I}{t})$ $= supp(I)+supp(t)$ we have $supp(t)\subseteq [-b-\bar{x},b]-supp(I)$, i.e. eq \eqref{loctsupp}.\\
To prove the condition in eq. \eqref{supphatt} note that eq. \eqref{locd} is equivalent in the Fourier domain to
\begin{equation}
\scal{D_{s}I}{t}=\scal{\widehat{D_{s}I}}{\hat{t}} = \frac{1}{s}\int d\omega\;\hat{I}\big(\frac{\omega}{s}\big)\hat{t}(\omega)\neq 0\;\;\forall s\in[\frac{\bar{s}}{S},S]
\end{equation}
The situation is depicted in Fig. \ref{FigInvdt} $b')$ for $S$ big enough: in this case in fact we can suppose the support of $\widehat{D_{\bar{s}/S}I}$ to be on an interval on the left of that of $supp(\hat{t})$ and $\widehat{D_{S}I}$ on the right; the condition $supp(\scal{\widehat{D_{s}I}}{\hat{t}})\subseteq [\bar{s}/S,S]$ is in this case equivalent to
\begin{equation}
\omega_{M}^{I}\frac{\bar{s}}{S}<\omega^{t}_{m},\;\;\omega_{M}^{t}<\omega^{I}_{m}S
\end{equation}
which gives
\begin{equation}\label{suppht}
\Delta^{*}_{t}=\max (\Delta_{t})\equiv \max\Big( \frac{\omega^{t}_{M}-\omega_{m}^{t}}{2}\Big)= S\omega_{m}^{I}-\omega^{I}_{M}\frac{\bar{s}}{S}
\end{equation}
and therefore eq. \eqref{supphatt}.
Note that for some $s\in[\bar{s}/S,S]$ the condition that the Fourier supports are disjoint is only sufficient and not necessary for the dot product to be zero since cancelations can occur. However we can repeat the reasoning done for the translation case and ask for $\scal{\widehat{D_{s}I}}{\hat{t}}=0$ on a continuous interval of scales.Q.E.D.\\
\begin{figure}\centering
\includegraphics[width= 8cm]{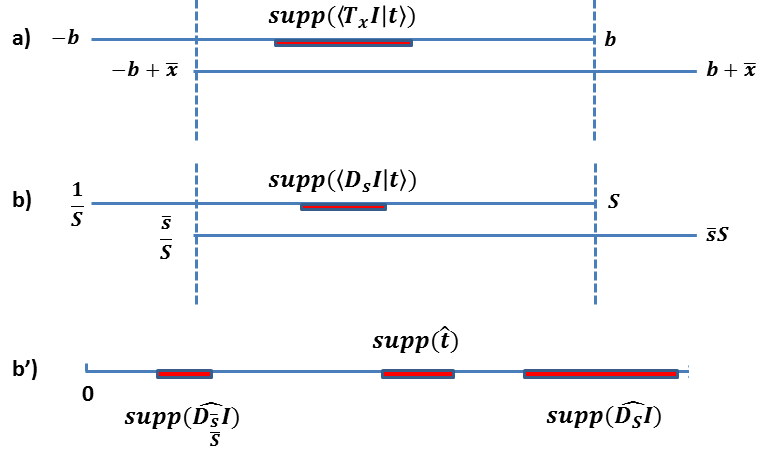}
\caption{$a),b)$: if the support of the dot product between the
    image and the template is contained in the intersection between
    the pooling range and the group translated (a) or dilated (b)
    pooling range the signature is invariant. In frequency, condition
    b) becomes b'): when the Fourier supports of the dilated image and the template do not intersect their dot product is zero.
\label{FigInvdt}}
\end{figure}
The results above lead to a statement connecting {\it invariance} with
{\it localization} of the templates:
\vspace{0.1in}
\begin{theorem}\label{invariance_implies_localization}
  Maximum translation invariance implies a template with minimum
  support in the space domain ($x$); maximum scale invariance implies
  a template with minimum support in the Fourier domain ($\omega$).
\end{theorem}
{\bf Proof:}\\
\noindent We illustrate the statement of the theorem with a simple example. In the case of translations suppose, e.g., $supp(I)=[-b',b'],\;supp(t)=[-a,a]$, $a\leq b' \leq b$. Eq. \eqref{loctsupp} reads
$$
[-a,a]\subseteq [-b+\bar{x}+b',b-b']
$$
which gives the condition $-a\geq -b+b'+\bar{x}$,
i.e. $\bar{x}^{max}=b-b'-a$; thus, for any fixed $b,b'$ the smaller
the template support, $2a$, in space, the greater is translation invariance.\\
Similarly, in the case of dilations, increasing the range of
invariance $[1,\bar{s}],\;\bar{s}>1$ implies a decrease in
the support of $\hat{t}$ as shown by eq. \eqref{suppht};
in fact noting that $|supp(\hat{t})|=2\Delta_{t}$ we have
$$
\frac{d|supp(\hat{t})|}{d\bar{s}} =  -\frac{2\omega^{I}_{M}}{S}<0
$$
i.e. the measure, $|\cdot|$, of the support of $\hat{t}$ is a decreasing function w.r.t. the measure of the invariance range $[1, \bar{s}]$. Q.E.D.\\
Because of the assumption of maximum possible support of all $I$ being
finite there is always localization  for any choice of
$I$ and $t$ under spatial shift. Of course if the localization support
is larger than the pooling range there is no invariance.\\
\noindent For a complex cell with pooling range $[-b, b]$ in space
only templates with self-localization smaller than the pooling range
make sense. An extreme case of self-localization is $t (x) =
\delta(x)$, corresponding to maximum localization of tuning of the
simple cells.

\subsection{Invariance, Localization and Wavelets}
The conditions equivalent to optimal translation and scale invariance --
maximum localization in space and frequency -- cannot be simultaneously
satisfied because of the classical {\it uncertainty principle}: if a
function $t(x)$ is essentially zero outside an interval of length
$\Delta x$ and its Fourier transform $\hat{I}(\omega)$ is essentially
zero outside an interval of length $\Delta \omega$ then
\begin{equation}
\Delta x \cdot \Delta \omega \ge 1.
\end{equation}
\noindent In other words a function and its Fourier transform cannot both be highly
concentrated. Interestingly for our setup the uncertainty principle
also applies to sequences\cite{donoho}.

It is well known  that the equality sign in the uncertainty principle
above is achieved by Gabor functions\cite{gabor} of the form
\begin{equation}
\psi_{x_0, \omega_0} (x)= e^{- \frac{(x-x_0)^2}{2 \sigma^2}} e^{i \omega_0 (x-x_0)},\;\sigma\in \R^{+},\;\omega_{0},x_{0}\in \R
\end{equation}
The uncertainty principle leads to the concept of ``optimal
  localization'' instead of exact localization. In a similar way, it
  is natural to relax our definition of strict invariance (e.g.
  $\mu_{n}^{k}(I) = \mu_{n}^{k}(g'I)$) and to introduce $\epsilon$-invariance as $|\mu_{n}^{k}(I) - \mu_{n}^{k}(g'I)| \leq
  \epsilon$. In particular if we suppose, e.g., the following localization condition
  \begin{equation}
  \scal{T_{x}I}{t}=e^{-\frac{x^{2}}{\sigma^{2}_{x}}},\;\;\;\scal{D_{s}I}{t}=e^{-\frac{s^{2}}{\sigma^{2}_{s}}},\;\sigma_{x},\sigma_{s}\in \R^{+}
  \end{equation}
  we have
\begin{eqnarray*}
 && |\mu_{n}^{k}(T_{\bar{x}}I)-\mu_{n}^{k}(I)| = \frac{1}{2}\sqrt{\sigma_{x}}\Big(\text{erf}\big([-b,b]\Delta[-b+\bar{x},b+\bar{x}]\big)\Big)\\
 && |\mu_{n}^{k}(D_{\bar{s}}I)-\mu_{n}^{k}(I)| = \frac{1}{2}\sqrt{\sigma_{s}}\Big(\text{erf}\big([-1/S,S]\Delta[\bar{s}/S,S\bar{s}]\big)\Big).
 \end{eqnarray*}
 where $\text{erf}$ is the error function.
The differences above, with an opportune choice of the localization ranges $\sigma_{s},\sigma_{x}$ can be made as small as wanted.\\
We end this paragraph by a \emph{conjecture}:
the optimal $\epsilon-$invariance is  satisfied by templates with non compact
support which decays exponentially such as a Gaussian or a Gabor
wavelet. We can then speak of {\it
optimal invariance} meaning ``optimal $\epsilon$-invariance''. The reasonings above lead to the
theorem:
\begin{theorem}\label{MainTheorem} Assume invariants are computed from
  pooling within a pooling window with a set of linear filters. Then  the
  optimal templates (e.g. filters) for maximum simultaneous
  invariance to translation and scale are Gabor
  functions
  \begin{equation}
  t(x) = e^{- \frac{x^2}{2 \sigma^2}}  e^{i \omega_0 x}.
  \end{equation}
\end{theorem}
\noindent {\bf Remarks}\\
\begin{enumerate}

\item  The Gabor function $\psi_{x_0, \omega_0} (x)$ corresponds
  to a {\it Heisenberg box} which has a $x$-spread $\sigma_x^2 = \int
  x^2 |\psi(x)| dx $ and a $\omega$ spread $\sigma_\omega^2 = \int
  \omega^2 |\hat{\psi}(\omega)| d \omega$ with area $\sigma_x \sigma_\omega$.
Gabor wavelets arise under the action on $\psi(x)$ of
the translation and scaling groups as follows. The function $\psi(x)$,
as defined, is zero-mean and normalized that is
\begin{equation}
\int \psi(x) dx= 0
\end{equation}
\noindent and
\begin{equation}
||\psi(x)||= 1.
\end{equation}
A family of Gabor wavelets is obtained by translating and scaling $\psi$:
\begin{equation}
\psi_{u,s}(x) = \frac{1}{s^{\frac{1}{2}}} \psi(\frac{x-u}{s}).
\end{equation}
Under certain conditions (in particular, the Heisenberg boxes
associated with each wavelet must together cover the space-frequency
plane) the Gabor wavelet family becomes a Gabor wavelet frame.

\item Optimal self-localization of the templates (which follows from
  localization), when valid simultaneously for space and scale, is also
  equivalent to Gabor wavelets. If they are a frame, full
  information can be preserved in an optimal quasi invariant way.

\item Note that the result proven in Theorem \ref{MainTheorem} is not
  related to that in \cite{Stevens2004}. While Theorem
  \ref{MainTheorem} shows that Gabor functions emerge from the
  requirement of maximal {\it invariance} for the complex cells -- a
  property of obvious computational importance -- the main result in
  \cite{Stevens2004} shows that (a) the wavelet transform (not
  necessarily Gabor) is covariant with the similitude group and (b)
  the wavelet transform follows from the requirement of covariance
  (rather than invariance) for the simple cells (see our definition of
  covariance in the next section). While (a) is well-known (b) is less
  so (but see p.41 in \cite{antoine}). Our result shows that Gabor
  functions emerge from the requirement of maximal invariance for the
  {\it complex} cells.
\end{enumerate}

\subsection{Approximate Invariance and Localization}

In the previous section we analyzed the relation between localization
and invariance in the case of group transformations. By relaxing
the requirement of exact invariance and exact localization we show how
the same strategy for computing invariants can still be applied even
in the case of non-group transformations if certain localization
properties of $\scal{TI}{t}$ holds, where $T$ is a smooth
transformation (to make it simple think to a transformation parametrized by one parameter).

We first notice that the localization condition of theorems
\ref{generalinv} and \ref{MainTheorem} -- when relaxed to approximate
localization -- takes the (e.g. for the $1D$ translations group supposing for simplicity that the supports of $I$ and $t$ are centered in zero) form
$\scal{I}{T_xt^{k}}< \delta \quad \forall x\;s.t.\; |x|>a$, where
$\delta$ is small in the order of $1/\sqrt{n}$ (where $n$ is the dimension of the space) and $\scal{I}{T_{x}t^{k}}\approx 1 \quad \forall x\;s.t.\;|x|<a$.\\
We call this property {\it sparsity of $I$ in the dictionary $t^k$
  under $G$.} This condition can be satisfied by templates that are
similar to images in the set and are sufficiently ``rich'' to be
incoherent for ``small'' transformations.  Note that from the
reasoning above the sparsity of $I$ in $t^k$ under
$G$ is expected to improve with increasing $n$ and with noise-like
encoding of $I$ and $t^k$ by the architecture.\\
Another important property of sparsity of $I$ in $t^k$ (in addition to
allowing local approximate invariance to arbitrary transformations,
see later) is {\it clutter-tolerance} in the sense that if $n_1,
n_2$ are additive uncorrelated spatial noisy clutter $\scal{I+n_1}{gt^k + n_2} \approx \scal{I}{g t}$.\\
Interestingly the {\it sparsity condition under the group} is related
to associative memories for instance of the holographic type\cite{Poggio1975},\cite{plate1991}. If the sparsity condition
holds only for $I=t^k$ and for very small set of $g\in G$, that is, it
has the form $\scal{I}{g t^{k}} = \delta(g) \delta_{I,t^{k}}$ it
implies strict memory-based recognition ( see non-interpolating
look-up table in the description of \cite{Poggio1990}) with inability
to generalize beyond stored templates or views.

While the first regime -- exact (or $\epsilon-$) invariance  for generic images,
yielding universal Gabor templates -- applies to the first layer of
the hierarchy, this second regime (sparsity) -- approximate invariance for a
class of images, yielding class-specific templates -- is important for
 dealing with non-group transformations at the top levels of the
 hierarchy where receptive fields may be as large as the visual field.

Several interesting transformations do not have the group
structure, for instance the change of expression of a face or
the change of pose of a body. We show here that  approximate invariance to
transformations that are not groups can be obtained if the approximate
localization condition above holds, and if the
transformation can be locally approximated by a linear transformation,
e.g. a combination of translations, rotations and non-homogeneous
scalings, which corresponds to a locally compact
group admitting a Haar measure.

Suppose, for simplicity, that the smooth transformation  $T$, at least twice differentiable, is parametrized by the parameter $r\in\R$.  We
 approximate its action on an image $I$ with a Taylor series (around e.g. $r=0$) as:
\begin{eqnarray}\label{linear}
T_{r}(I) &=& T_{0}(I)+\Big(\frac{dT}{dr}\Big)_{r=0}(I)r +R(I)\\
  &=& I+\Big(\frac{dT}{dr}\Big)_{r=0}(I)r +R(I)\nonumber\\
   &=& I+J^{I}(I)r +R(I) = [e+rJ^{I}](I)+R(I)\nonumber \\
   &=& L_{r}^{I}(I)+R(I)\nonumber
\end{eqnarray}
\noindent where $R(I)$ is the reminder, $e$ is the identity
operator, $J^{I}$ the Jacobian and $L^{I}_{r}= e+ J^{I}r$ is a linear
operator.\\  Let $R$ be the range of the parameter $r$ where we can
approximately neglect the remainder term $R(I)$. Let $L$ be the range of
the parameter $r$ where the scalar product $\scal{T_{r}I}{t}$ is
localized i.e.  $\scal{T_{r}I}{t}=0,\;\forall r\not\in L$. If $L\subseteq R$ we have
\begin{equation}
\scal{T_{r}I}{t} \approx  \scal{L^{I}_{r}I}{t},
\end{equation}
and we have  the following:
\vspace{0.05in}
\begin{proposition}
Let $I,t\in H$ a Hilbert space, $\eta_{n}:\R\rightarrow\R^{+}$ a set of bijective (positive) functions and $T$ a smooth transformation (at least twice differentiable) parametrized by $r\in\R$.  Let $L=supp(\scal{T_{r}I}{t})$, $P$ the pooling interval in the $r$ parameter and $R\subseteq \R$ defined as above. If $L\subseteq P \subseteq R $ and
$$
\scal{T_{r}I}{t}=0,\;\forall r\in \R/(T_{\bar{r}}P\cap P),\;\bar{r}\in \R
$$
then $\mu^{k}_{n}(T_{\bar{r}}I)=\mu^{k}_{n}(I)$.
\end{proposition}
\noindent
{\bf Proof:}\\
\noindent
We have, following the reasoning done in Theorem \ref{generalinv}
\begin{eqnarray*}
\mu^{k}_{n}(T_{\bar{r}}I)&=& \int_{P}dr\;\eta_{n}(\scal{T_{r}T_{\bar{r}}I}{t})= \int_{P}dr\;\eta_{n}(\scal{L^{I}_{r}L^{I}_{\bar{r}}I}{t})\\
                         &=& \int_{P}dr\;\eta_{n}(\scal{L^{I}_{r+\bar{r}}I}{t}) =\mu^{k}_{n}(I)
\end{eqnarray*}
where the last equality is true if  $\scal{T_{r}I}{t}=\scal{L^{I}_{r}I}{t}=0,\;\forall r\in \R/(T_{\bar{r}}P\cap P)$.
Q.E.D.\\\\
As an example, consider the transformation induced on the image plane by
rotation in depth of a face: it can be decomposed into piecewise linear approximations
around a small number of key templates, each one corresponding to a specific
$3D$ rotation of a template face. Each key template corresponds to a
complex cell containing as (simple cells) a number of observed
transformations of the key template within a small range of
rotations. Each key template corresponds to a different signature
which is invariant only for rotations around its center. Notice that the
form of the linear approximation or the number of key templates needed
does not affect the algorithm or its implementation. The templates
learned are used in the standard dot-product-and-pooling module. The
choice of the key templates -- each one corresponding to a complex
cell, and thus to a signature component -- is not critical, as long as there are
enough of them. For one parameter groups, the key templates correspond
to the knots of a piecewise linear spline approximation. Optimal
placement of the centers -- if desired -- is a separate problem that we leave aside for now.

\noindent {\bf Summary of the argument:}
Different transformations can be classified in terms
of invariance and localization.

{\it Compact Groups:} consider the case of a compact group
transformation such as rotation in the image plane. A complex cell is
invariant when pooling over all the templates which span the full
group $\theta\in [-\pi,+\pi]$. In this case there is no restriction on
which images can be used as templates: any template yields perfect
invariance over the whole range of transformations (apart from mild
regularity assumptions) and a single complex cell pooling over all
templates can provide a globally invariant signature.

{\it Locally Compact Groups and Partially Observable Compact Groups:}
consider now the POG situation in which the pooling is over a subset
of the group: (the POG case always applies to Locally Compact  groups (LCG) such as
translations). As shown before, a complex cell is
partially invariant {\bf if} the value of the dot-product between a
template and its shifted template under the group falls to zero fast
enough with the size of the shift relative to the extent of
pooling.

In the POG and LCG case, such partial invariance holds over a
restricted range of transformations if the templates and the inputs
have a {\it localization} property that implies wavelets for
transformations that include translation and scaling.

{\it General (non-group) transformations:} consider  the case of a smooth transformation which may not be a group. Smoothness
implies that the transformation can be approximated by piecewise linear
transformations, each centered around a template (the local linear operator
corresponds to the first term of the Taylor series expansion around
the chosen template). Assume -- as in the POG case -- a special form
of {\it sparsity} -- the dot-product between the template and its
transformation fall to zero with increasing size of the transformation.
 Assume also that the templates
transform as the input image. For instance, the transformation induced
on the image plane by rotation in depth of a face may have piecewise
linear approximations around a small number of key templates
corresponding to a small number of rotations of a given template face
(say at $\pm 30^o, \pm90^o, \pm 120^o$). Each key template and its
transformed templates within a range of rotations corresponds to
complex cells (centered in $\pm 30^o, \pm90^o, \pm 120^o$). Each key
template, e.g. complex cell, corresponds to a different signature which
is invariant only for that part of rotation.  The strongest hypothesis
is that there exist input images that are sparse w.r.t. templates of the
same class -- these are the images for which local invariance
holds.\\
\noindent {\bf Remarks}:
\vspace{0.05in}
\begin{enumerate}
\item We are interested in two main cases of POG invariance:

\begin{itemize}
\item partial invariance {\it simultaneously} to translations in $x,y$, scaling and
  possibly rotation in the image plane. This should apply to  ``generic''
  images. The signatures should ideally preserve full,
  locally invariant information. This first regime is ideal for the
  first layers of the multilayer network and may be related to Mallat's
  scattering transform, \cite{MallatGroupInvariantMain}. We call the sufficient condition for
  LCG invariance here, {\it localization}, and in particular, in the case of translation (scale) group {\it
    self-localization} given by Equation \eqref{SD}.

\item partial invariance to linear transformations for a subset of
  all images. This second regime applies to high-level modules in the
  multilayer network specialized for specific classes of objects and
  non-group transformations. The condition that is sufficient here for
  LCG invariance is given by Theorem \ref{generalinv} which applies only
  to a specific class of $I$. We prefer to call it {\it sparsity} of the images with respect to a set of
  templates.

\end{itemize}

\item For classes of images that are sparse with respect to
 a set of templates, the localization condition does not imply wavelets. Instead it implies templates that are
\begin{itemize}
\item similar to a class of images so that $\scal{I}{g_0 t^{k}} \approx1$ for some $g_{0}\in G$ and
\item  complex enough to be  ``noise-like'' in the sense that $\scal{I}{gt^{k}} \approx 0$ for $g\neq g_{0}$.
\end{itemize}

\item Templates must transform similarly to the input for approximate invariance
  to hold.  This corresponds to the assumption of a class-specific
  module and of a {\it nice object class} \cite{poggio1992recognition,Leibo2011b}.
\item For the localization property to hold, the image must be similar
  to the key template or contain it as a diagnostic feature (a
  sparsity property). It must
  be also quasi-orthogonal (highly localized) under the action of the local group.
\item For a general, non-group, transformation it may be impossible to obtain
  invariance over the full range with a single signature; in general
  several are needed.

\item It would be desirable to derive a formal characterization of the error
in local invariance by using the standard module of
dot-product-and-pooling, equivalent to a complex cell. The above
arguments provide the outline of a proof based on local
linear approximation of the transformation and on the fact that a
local linear transformation is a LCG.

\end{enumerate}

\vspace{0.3in}

\section*{3. Hierarchical Architectures}

So far we have studied the invariance, uniqueness and stability
properties of signatures, both in the case when a whole group of
transformations is observable (see~\eqref{disc_meas}
and~\eqref{cont_meas}), and in the case in which it is only partially
observable (see~\eqref{POG_disc_meas} and~\eqref{mu}).  We now discuss
how the above ideas can be iterated to define a multilayer
architecture.  Consider first the case when $G$ is finite. Given a
subset $G_0\subset G$, we can associate a {\em window} $gG_0$ to each
$g\in G$.  Then, we can use definition~\eqref{POG_disc_meas} to define
for each window a signature $\Sigma(I)(g)$ given by the measurements,
$$\mu^{k}_{n}(I)( g)=\frac{1}{|G_0|}\sum_{\bar g\in {g}G_0}\eta_n\Big(\scal{I}{\bar{g}t^{k}}\Big).$$
We will keep this form as the definition of signature.
For fixed $n,k$, a set of measurements corresponding to different
windows can be seen as a $|G|$ dimensional vector.  A signature
$\Sigma(I)$ for the whole image is obtained as {\em a signature of
  signatures}, that is, a collection of signatures $(\Sigma(I)(g_1),
\dots,\Sigma(I)(g_{|G|})$
associated to each window.\\  Since we assume that the output of each
module is made zero-mean and normalized before further processing at
the next layer, {\it conservation of information
from one layer to the next requires saving the mean and the norm} at
the output of each module at each level of the hierarchy.\\
We {\em conjecture} that the neural image at the first layer is uniquely represented by the final signature at the top of the hierarchy and the means and norms at each layer.\\
The above discussion can be easily extended to continuous (locally
compact) groups considering,
$$
\mu^{k}_{n}(I)( g)= \frac{1}{V_0}\int_{gG_0}  d\bar{g}\eta_n\Big(\scal{I}{\bar{g}t^{k}}\Big), \quad  V_0=\int_{G_0} d\bar{g},
$$
where, for fixed $n,k$, $\mu^{k}_{n}(I):G \to\reals$ can now be seen
as a function on the group.  In particular, if we denote by $K_0:G\to
\reals$ the indicator function on $G_0$, then we can write
$$
\mu^{k}_{n}(I)( g)= \frac{1}{V_0}\int_{G} d\bar{g} K_0(\bar{g}^{-1} g) \eta_n\Big(\scal{I}{\bar{g}t^{k}}\Big).
$$
The signature for an image can again be seen as a collection of
signatures corresponding to different windows, but in this case it is
a function $\Sigma(I):G\to \reals^{NK}$, where $\Sigma(I)(g)\in
\reals^{NK}$, is a signature corresponding to the window $G_0$
``centered'' at $g\in G$.

The above construction can be iterated  to define  a hierarchy of  signatures.
Consider a sequence $G_{1}\subset G_2\subset,  \dots, \subset G_{L}= G$.
For $h: G\to \reals^p$, $p\in {\mathbb N}$ with an abuse of notation we let $gh(\bar g)=h(g^{-1}\bar g)$.
Then we can consider the following construction.

We call \textit{complex cell operator} at layer $\ell$ the operator
that maps an image $I\in \XX$ to a function $\mu_\ell(I):G\to
\reals^{NK}$ where
\begin{equation}\label{h}
\mu^{n,k}_{\ell}(I)(g) = \frac{1}{|G_{\ell}|}\sum_{\bar{g}\in gG_{\ell}}\eta_n\left (\nu^k_{\ell}(I)(\bar g)\right),
\end{equation}
and \textit{simple cell operator} at layer $\ell$ the operator that
maps an image $I\in \XX$ to a function $\nu_\ell(I):G\to \reals^{K}$
\begin{equation}\label{sl}
\nu^{k}_{\ell}(I)(g) = \scal{\mu_{\ell-1}(I)}{gt^k_{\ell}}
\end{equation}
with $t^{k}_{\ell}$ the $k^{th}$ template at layer $\ell$ and
$\mu_{0}(I)= I$.  Several comments are in order:
\begin{itemize}
\item beside the first layer, the inner product defining the simple
  cell operator is that in $L^2(G)=\{h:G\to \reals^{NK},~|~\int
  dg|h(g)|^2<\infty\}$;
\item The index $\ell$ corresponds to different layers, corresponding
  to different subsets $G_\ell$.
\item At each layer a (finite) set of templates ${\cal
    T}_\ell=(t_\ell^1,\dots,t_\ell^{K}) \subset L^2(G)$ (${\cal T}_0
  \subset \XX$) is assumed to be available. For simplicity, in the
  above discussion we have assumed that $|{\cal T}_\ell | = K$, for
  all $\ell=1, \dots,L$.  The templates at layer $\ell$ can be thought
  of as \textit{compactly supported functions}, with support much
  smaller than the corresponding set $G_\ell$. Typically templates can
  be seen as image patches in the space of complex operator responses,
  that is $t_\ell=\mu_{\ell-1}(\bar{t})$ for some $\bar{t} \in \XX$.
\item Similarly we have assumed that the number of non linearities
  $\eta_n$, considered at every layer, is the same.
\end{itemize}

Following the above discussion, the extension to continuous (locally
compact) groups is straightforward. We collect it in the following
definition.
\vspace{0.1in}
\begin{definition*}[Simple and complex response]\label{deep}
  For $\ell=1, \dots, L$, let ${\cal T}_\ell=(t_\ell^1,\dots,t_\ell^K)
  \subset L^2(G)$ (and ${\cal T}_0 \subset \XX$) be a sequence of
  template sets.  The complex cell operator at layer $\ell$ maps an
  image $I\in \XX$ to a function $\mu_\ell(I):G\to \reals^{NK}$; in
  components
\begin{equation}\label{cl}
\mu^{n,k}_{\ell}(I)(g) = \frac{1}{V_{\ell}}
\int d\bar{g} K_\ell(\bar{g}^{-1}g) \eta_n\left (\nu^k_{\ell}(I)(\bar g)\right),\;g\in G
\end{equation}
where $K_{\ell}$ is the indicator function on $G_{\ell}$, $V_{\ell}=\int_{G_{\ell}}d\bar{g}$ and where
\begin{equation}\label{sl2}
\nu^{k}_{\ell}(I)(g) = \scal{\mu_{\ell-1}(I)}{gt^k_\ell},\quad g\in G
\end{equation}
($\mu_{0}(I)= I$) is the simple cell operator at layer $\ell$ that maps an image $I\in \XX$ to a function $\nu_\ell(I):G\to \reals^{K}$.
\end{definition*}
\noindent {\bf Remark}
Note that eq. \eqref{cl} can be written as:
\begin{equation}\label{conv}
\mu^{n,k}_{\ell}(I) = K_{\ell}*\eta_{n}(\nu^{k}_{\ell}(I))
\end{equation}
where $*$ is the group convolution.\\

In the following we study the properties of the complex response, in particular

\subsection{Property 1: covariance}
\begin{figure}\centering
\includegraphics[width= 8cm]{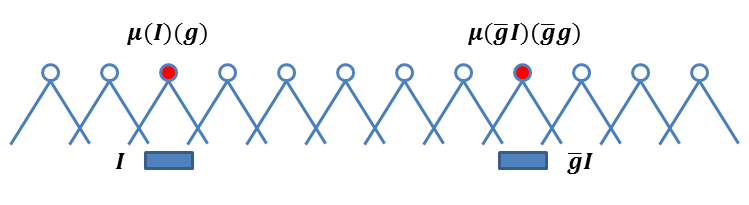}
\caption{Covariance: the response for an image $I$ at position $g$ is equal to the response of the group shifted image at the shifted position.\label{C}}
\end{figure}

We call the map $\Sigma$  covariant under $G$ iff
$$
\Sigma(gI)=g^{-1}\Sigma(I),\quad \forall g\in G, I \in\XX
$$,
where the action of $g^{-1}$ is intended on the representation space
$L^2(G)$ and that of $g$ on the image space $L^{2}(\R^2)$. Practically since we are only taking into account of the distribution of the values of
$\scal{\mu(I)}{\mu(t^{k})}$ we can ignore this technical detail being the definition of covariance equivalent to the statement
$\scal{\mu(gI)}{\mu(t^{k})}=\scal{\mu(I)}{\mu(g^{-1}t^{k})}$ where the transformation is always acting on the image space.
In the following we show the covariance property for the $\mu^{n,k}_{1}$
response (see Fig. \ref{C}). An inductive reasoning then can be applied for higher order
responses. We assume that the architecture is isotropic in the
relevant covariance dimension (this implies that all the modules in
each layer should be identical with identical templates) and that
there is a continuum of modules in each layer.

\begin{proposition}\label{covgen}
Let $G$ a locally compact group and $\bar{g}\in G$. Let $\mu^{n,k}_{1}$ as defined in eq. \ref{cl}. Then $\mu^{n,k}_{1}(\tilde{g}I)(g) = \mu^{n,k}_{1}(I)(\tilde{g}^{-1}g),\;\forall \tilde{g}\in G$.
\end{proposition}
\noindent
{\bf Proof:}\\
Using the definition  \ref{cl} we have
\begin{eqnarray*}
\mu^{n,k}_{1}(\tilde{g}I)(g) &=& \frac{1}{V_{1}}
\int_{G} d\bar{g} K_1(\bar{g}^{-1}g) \eta_n\left (\scal{\tilde{g}I}{\bar{g}t^{k}}\right)\\
&=& \frac{1}{V_{1}}\int_{G} d\bar{g} K_1(\bar{g}^{-1}g) \eta_n\left (\scal{I}{\tilde{g}^{-1}\bar{g}t^{k}}\right)\\
&=& \frac{1}{V_{1}}\int_{G} d\hat{g} K_1(\hat{g}^{-1}\tilde{g}^{-1}g) \eta_n\left (\scal{I}{\hat{g}t^{k}}\right)\\
&=& \mu^{n,k}_{1}(I)(\tilde{g}^{-1}g)
\end{eqnarray*}
where in the third line we used the change of variable
$\hat{g}=\tilde{g}^{-1}\bar{g}$ and the invariance of the Haar
measure. Q.E.D.\\
\noindent {\bf Remarks}
\begin{enumerate}
\item The covariance property described in proposition \ref{covgen} can be stated equivalently as $\mu_{1}^{n,k}(I)(g)= \mu_{1}^{n,k}(\bar{g}I)(\bar{g}g)$. This last expression has a more intuitive meaning as shown in Fig. \ref{C}.

\item The covariance property described in proposition \ref{covgen} holds both for abelian and non-abelian groups. However the group average on templates transformations in definition of eq. \ref{cl} is crucial. In fact, if we define the signature averaging on the images we do not have a covariant response:
\begin{eqnarray*}
\mu^{n,k}_{1}(\tilde{g}I)(g) &=& \frac{1}{V_{1}}
\int_{G} d\bar{g} K_1(\bar{g}^{-1}g) \eta_n\left (\scal{\bar{g}\tilde{g}I}{t^{k}}\right)\\
&=& \int_{G} d\hat{g} K_1(\tilde{g}\hat{g}^{-1}g) \eta_n\left (\scal{\hat{g}I}{t^{k}}\right)
\end{eqnarray*}
where in the second line we used the change of variable $\hat{g}=\tilde{g}^{-1}\bar{g}$ and the invariance of the Haar measure.
The last expression cannot be written as $\mu^{n,k}_{1}(I)(g'g)$
for any $g'\in G$.\\

\item With respect to the range of invariance, the following property
  holds for multilayer architectures in which the output of a layer
  is defined as covariant if it transforms in the same way as the
  input: for a given transformation of an image or part of it, the
  signature from complex cells at a certain layer is either invariant
  or covariant with respect to the group of transformations; if it is
  covariant there will be a higher layer in the network at which it is
  invariant (more formal details are given in Theorem \ref{WPsm}),
  assuming that the image is contained in the visual field.
    This property predicts a {\em stratification} of ranges
  of invariance in the ventral stream: invariances should appear in a
  sequential order meaning that smaller transformations will be
  invariant before larger ones, in earlier layers of the hierarchy\cite{Isik2013}.
\end{enumerate}

\subsection{Property 2: partial and global invariance (whole and parts)}
\begin{figure}\centering
\includegraphics[width= 8cm]{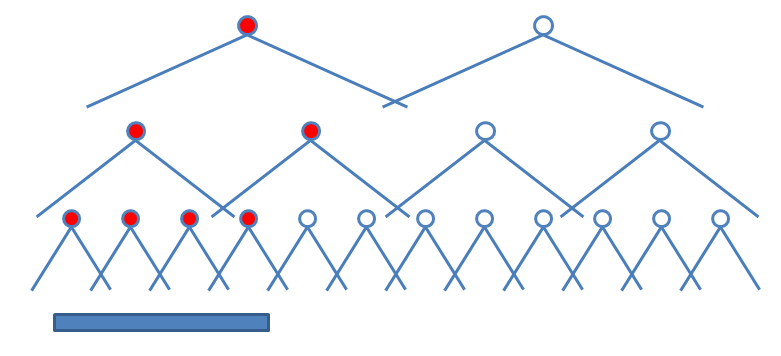}
\caption{An image $I$ with a finite support may or may not be
fully included in the receptive field of a single complex cell at
layer $n$ (more in general the transformed image may not be included
in the pooling range of the complex cell). However there will be a
higher  layer such that the support of its neural response is included in the pooling range of a single complex cell.\label{WP}}
\end{figure}

We now find the conditions under which  the functions $\mu_\ell$ are locally invariant, i.e. invariant within the restricted range of the
pooling. We further prove that the range of invariance increases from layer to layer in the hierarchical architecture. The fact that  for an image, in general, no more global invariance is guaranteed  allows, as we will see, a novel definition of ``parts'' of an image.\\
The local invariance conditions are a simple reformulation of Theorem \ref{generalinv} in the context of a hierarchical architecture. In the following, for sake of simplicity, we suppose that at each layer we only have a template $t$ and a non linear function $\eta$.\\
\vspace{0.05in}
\begin{proposition}{\bf Localization and  Invariance: hierarchy.}\setcounter{proof}{7}
\label{Tinv}
Let $I,t\in H$ a Hilbert space, $\eta:\R\rightarrow\R^{+}$ a bijective (positive) function and $G$ a locally compact group. Let $G_{\ell}\subseteq G$ and suppose $supp(\scal{g\mu_{\ell-1}(I)}{t})\subseteq G_{\ell}$. Then for any given $\bar{g}\in G$
\begin{eqnarray}\label{loch}
&&\scal{g \mu_{\ell-1}(I)}{t}=0,\; g\in G/(G_{\ell}\cap\bar{g}G_{\ell})\nonumber\\
&&\textrm{or equivalently}\;\;\;\;\;\;\;\;\;\;\;\;\;\;\;\;\;\;\;\;\;\;\;\;\;\;\;\;\;\;\;\;\;\;\Rightarrow\;\mu_{\ell}(I) = \mu_{\ell}(\bar{g}I)\nonumber\\
&&\scal{g \mu_{\ell-1}(I)}{t}\neq 0,\; g\in G_{\ell}\cap\bar{g}G_{\ell}.
\end{eqnarray}
\end{proposition}
The proof follows the reasoning done in Theorem \ref{generalinv} (and the following discussion for the translation and scale transformations)  with $\im$ substituted by $\mu_{\ell-1}(\im)$ using the covariance property
$\mu_{\ell-1}(gI)=g\mu_{\ell-1}(I)$. Q.E.D.\\
\noindent
We can give now a formal definition of \emph{object part} as the
subset of the signal $\im$ whose complex response, at layer $\ell$, is
invariant
under transformations in the range of the pooling at that layer.\\
\noindent
This definition is consistent since the invariance is increasing from
layer to layer (as formally proved below) therefore allowing bigger
and bigger parts.  Consequently for each transformation there will
exists a layer $\bar{\ell}$ such that any signal subset will be a part
at that layer.\\
We can now state the following:
\begin{theorem} {\bf Whole and parts.} \label{WPsm} Let $\im\in\XX$ (an
  image or a subset of it) and $\mu_\ell$ the complex response
  at layer $\ell$.  Let $G_{0}\subset \cdots \subset
  G_{\ell}\subset\cdots\subset G_{L}=G$ a set of nested
  subsets of the group $G$.  Suppose $\eta$ is a bijective (positive) function and that the template $t$ and the
  complex response at each layer has finite support.  Then $\forall
  \bar{g}\in G$, $\mu_\ell(I)$ is invariant for some
  $\ell=\bar{\ell}$, i.e.
$$
\mu_{m}(\bar{g}\im)=\mu_{m}(\im),\;\;
\exists \;\bar{\ell}\;\;s.t.\;\;\forall m\geq\bar{\ell}.
$$
\end{theorem}
The proof follows from the observation that the pooling range over the group is
a bigger and bigger subset of $G$ with growing layer number, in other words, there exists a layer such that
the image and its transformations are within the pooling range at that
layer (see Fig. \ref{WP}). This is clear since for any $\bar{g}\in G$ the nested sequence
$$
G_{0}\cap \bar{g}G_{0}\subseteq...\subseteq G_{\ell}\cap \bar{g}G_{\ell}\subseteq ...\subseteq G_{L}\cap \bar{g}G_{L}=G.
$$
will include, for some $\bar{l}$, a set $G_{\bar{\ell}}\cap \bar{g}G_{\bar{\ell}}$  such that
$$
\scal{g\mu_{\bar{\ell}-1}(I)}{t}\neq 0 \;\; g\in G_{\bar{\ell}}\cap \bar{g}G_{\bar{\ell}}
$$
being $supp(\scal{g\mu_{\bar{\ell}-1}(I)}{t})\subseteq G$.

\subsection{Property 3: stability}
Using the definition of stability given in \eqref{Stab}, we can
formulate the following theorem characterizing stability for the
complex response:
\vspace{0.1in}
\begin{theorem} {\bf Stability.}  \label{STsm} Let $\im, \im'\in\XX$ and
  $\mu_\ell$ the complex response at layer $l$. Let the
  nonlinearity $\eta$ a Lipschitz function with Lipschitz constant
  $L_{\eta}\leq 1$. Then
\begin{equation}\label{eqST}
\nor{\mu_\ell(\im)-\mu_\ell(I')} \leq \nor{I-I'}_{H},\;\forall\;\ell,\;\forall\; I,I'\in \XX.
\end{equation}
where $\nor{I-I'}_H= min_{g,g'\in G_{\ell}}\nor{gI-g'I'}_2$
\end{theorem}
The proof follows from a repeated application of the reasoning done in Theorem \ref{St}. See details in \cite{MM2013}.

\subsection{Comparison with stability  as in \cite{MallatGroupInvariantMain}}
 The same definition of stability we use (Lipschitz continuity)  was recently given by
\cite{MallatGroupInvariantMain}, in a related context.
Let $I,I'\in L^{2}(\reals^{2})$ and $\Phi:L^{2}(\reals^{2})\to
L^{2}(\reals^{2})$ a representation. $\Phi$ is stable if it is
Lipschitz continuous with Lipschitz constant $L \leq 1$, i.e., is a
non expansive map:
\begin{equation}\label{StabilityMallat}
\nor{\Phi(I)-\Phi(I')}_{2}\leq \nor{I-I'}_{2},\;\;\forall\;I,I'\in L^{2}(\reals^{2}).
\end{equation}
In particular in \cite{MallatGroupInvariantMain} the author is
interested in stability of group invariant scattering representations to the action of small diffeomorphisms close
to translations.  Consider transformations of the form
$I'(\bf{x})=L_{\tau}I(\bf{x})=I(\bf{x}-\tau(\bf{x}))$ (which can be
though as small diffeomorphic transformations close to translations
implemented by a displacement field $\tau:\reals^{2}\to
\reals^{2}$). A translation invariant operator $\Phi$ is said to be
Lipschitz continuous to the action of a $C^2(\R^2)$ diffeomorphisms if
for any compact $\Omega\subseteq\R^2$ there exists $C$ such that for all
$I\in L^2(\R^2)$ supported in $\Omega\subseteq \R^2$ and $\tau\in
C^2(\R^2)$
\begin{eqnarray}\label{MallatS}
&&\nor{\Phi(I)-\Phi(L_{\tau}I)}_{2} \leq\\
&&\leq C \nor{I}_{2}\Big(sup_{{\bf x}\in \reals^{2}}|\nabla\tau({\bf x})| + sup_{{\bf x}\in \reals^{2}}|H\tau({\bf x})|  \Big)\nonumber
\end{eqnarray}
where $H$ is the Hessian and $C$ a positive constant.\\
Condition \eqref{MallatS} is a different condition then that in eq. \eqref{eqST} since it gives a Lipschitz bound for a diffeomorphic transformation at each layer of the scattering representation.\\
Our approach differs in the assumption that small (close to identity) diffeomorphic transformations can be well approximated, at the first layer, as locally affine transformations or, in the limit, as local translations which therefore falls in the POG case. This assumption is substantiated by the  following reasoning in which any smooth transformation is seen as parametrized by the parameter $t$ (the $r$ parameter of the $T_{r}$ transformation in section 2), which can be thought,e.g., as time.\\
Let $T\subseteq \mathbb{R}$ be a bounded interval and $\Omega
\subseteq\mathbb{R}^{N}$ an open set and let $\Phi =\left( \Phi
  _{1},...,\Phi _{N}\right):T\times \Omega \rightarrow\mathbb{R}^{N}$
be $\mathcal{C}_{2}$ (twice differentiable), where $\Phi \left( 0,.\right) $ is the identity
map.  Here $\mathbb{R}^{N}$ is assumed to model the image plane,
intuitively we should take $N=2$, but general values of $N$ allow our
result to apply in subsequent, more complex processing stages, for
example continuous wavelet expansions, where the image is also
parameterized in scale and orientation, in which case we should take
$N=4$.  We write $\left( t,x\right) $ for points in $T\times \Omega $,
and interpret $\Phi \left( t,x\right) $ as the position in the image
at time $t$ of an observed surface feature which is mapped to $x=\Phi
\left( 0,x\right) $ at time zero. The map $\Phi $ results from the
(not necessarily rigid) motions of the observed object, the motions of
the observer and the properties of the imaging apparatus. The implicit
assumption here is that no surface features which are visible in
$\Omega $ at time zero are lost within the time interval $T$. The
assumption that $\Phi $ is twice differentiable reflects assumed
smoothness properties of the surface manifold, the fact that object
and observer are assumed massive, and corresponding smoothness
properties of the imaging apparatus, including eventual
processing.\\
\noindent
Now consider a closed ball $B\subset \Omega $ of radius $\delta >0$ which
models the aperture of observation. We may assume $B$ to be centered at
zero, and we may equally take the time of observation to be $t_{0}=0\in T$.
Let
\[
K_{t}=\sup_{\left( t,x\right) \in T\times B}\left\Vert \frac{\partial ^{2}}{
\partial t^{2}}\Phi \left( t,x\right) \right\Vert _{\mathbb{R}^{N}},\;K_{x}=\sup_{x\in B}\left\Vert \frac{\partial
^{2}}{\partial
x\partial t}\Phi \left( 0,x\right) \right\Vert _{\mathbb{R}^{N\times N}}.
\]
Here $\left( \partial /\partial x\right) $ is the spatial gradient in $\mathbb{R}
^{M}$, so that the last expression is spelled out as
\[
K_{x}=\sup_{x\in B}\left( \sum_{l=1}^{N}\sum_{i=1}^{N}\left( \frac{\partial^{2}}{\partial x_{i}\partial t}\Phi _{l}\left(
0,x\right) \right)^{2}\right) ^{1/2}.
\]
Of course, by compactness of $T\times B$ and the $\mathcal{C}_{2}$%
-assumption, both $K_{t}$ and $K_{x}$ are finite. The following
theorem is due to Maurer and Poggio:

\begin{theorem}\label{localA}
There exists $V\in\mathbb{R}^{N}$ such that for all $\left( t,x\right) \in T\times B$
\[
\left\Vert \Phi \left( t,x\right) -\left[ x+tV\right] \right\Vert _{\mathbb{R}^{N}}\leq K_{x}\delta \left\vert t\right\vert
+K_{t}\frac{t^{2}}{2}.
\]
\end{theorem}

\noindent The proof reveals this to be just a special case of
Taylor's theorem.\\
\noindent
{\bf Proof:} Denote $V\left( t,x\right) =\left( V_{1},...,V_{l}\right) \left( t,x\right)
=\left( \partial /\partial t\right) \Phi \left( t,x\right) $, $\dot{V}\left(
t,x\right) =\left( \dot{V}_{1},...,\dot{V}_{l}\right) \left( t,x\right)
=\left( \partial ^{2}/\partial t^{2}\right) \Phi \left( t,x\right) $, and
set $V:=V\left( 0,0\right) $. For $s\in \left[ 0,1\right] $ we have with
Cauchy-Schwartz%
\begin{eqnarray*}
\left\Vert \frac{d}{ds}V\left( 0,sx\right) \right\Vert _{\mathbb{R}
^{N}}^{2}&=&\sum_{l=1}^{N}\sum_{i=1}^{N}\left( \left( \frac{\partial ^{2}}{%
\partial x_{i}\partial t}\Phi _{l}\left( 0,sx\right) \right) x_{i}\right)
^{2}\\
&\leq& K_{x}^{2}\left\Vert x\right\Vert ^{2}\leq K_{x}^{2}\delta ^{2},
\end{eqnarray*}
whence
\begin{eqnarray*}
&&\left\Vert \Phi \left( t,x\right) -\left[ x+tV\right] \right\Vert  \\
&=&\left\Vert \int_{0}^{t}V\left( s,x\right) ds-tV\left( 0,0\right)
\right\Vert  \\
&=&\left\Vert \int_{0}^{t}\left[ \int_{0}^{s}\dot{V}\left( r,x\right)
dr+V\left( 0,x\right) \right] ds-tV\left( 0,0\right) \right\Vert  \\
&=&\left\Vert \int_{0}^{t}\int_{0}^{s}\frac{\partial ^{2}}{\partial t^{2}}%
\Phi \left( r,x\right) drds+t\int_{0}^{1}\frac{d}{ds}V\left( 0,sx\right)
ds\right\Vert  \\
&\leq &\int_{0}^{t}\int_{0}^{s}\left\Vert \frac{\partial ^{2}}{\partial t^{2}%
}\Phi \left( r,x\right) \right\Vert drds+\left\vert t\right\vert
\int_{0}^{1}\left\Vert \frac{d}{ds}V\left( 0,sx\right) \right\Vert ds \\
&\leq &K_{t}\frac{t^{2}}{2}+K_{x}\left\vert t\right\vert \delta .
\end{eqnarray*}
Q.E.D.\\
Of course we are more interested in the visible features themselves, than in
the underlying point transformation. If $I:
\mathbb{R}^{N}\rightarrow\mathbb{R}$ represents these features, for example as a spatial distribution of
gray values observed at time $t=0$, then we would like to estimate the evolved
image $I\left( \Phi \left( t,x\right) \right) $ by a translate $I\left(
x+tV\right) $ of the original $I$. It is clear that this is possible only
under some regularity assumption on $I$. The simplest one is that $I$ is
globally Lipschitz. We immediately obtain the following

\begin{corollary}
Under the above assumptions suppose that $I:\mathbb{R}^{N}\rightarrow\mathbb{R}$ satisfies
\[
\left\vert I\left( x\right) -I\left( y\right) \right\vert \leq c\left\Vert
x-y\right\Vert
\]
for some $c>0$ and all $x,y\in\mathbb{R}^{N}$. Then there exists $V\in\mathbb{R}^{N}$ such that for all $\left( t,x\right)
\in I\times B$%
\[
\left\vert I\left( \Phi \left( t,x\right) \right) -I\left( x+tV\right)
\right\vert \leq c\left( K_{x}\left\vert t\right\vert \delta +K_{t}\frac{%
t^{2}}{2}\right) .
\]
\end{corollary}
\vspace{0.05in}
\noindent Theorem \ref{localA} and corollary 14 gives a precise mathematical motivation for the assumption that any sufficiently smooth (at least twice differentiable) transformation can be approximated in an enough small compact set with a group transformation (e.g. translation), thus allowing, based on  eq. \ref{Stab}, stability w.r.t. small diffeomorphic transformations.

\subsection{Approximate Factorization: hierarchy}
In the first version of \cite{MM2013} we conjectured that a
signature invariant to a group of transformations could be obtained by
factorizing in successive layers the computation of signatures
invariant to a subgroup of the transformations (e.g. the subgroup of
translations of the affine group) and then adding the invariance
w.r.t.  another subgroup (e.g. rotations). While factorization of
invariance ranges is possible in a hierarchical architecture (theorem \ref{WPsm}), it can
be shown that in general the factorization in successive layers for
instance of invariance to translation followed by invariance to
rotation (by subgroups) is impossible\cite{MM2013}.\\
However, approximate factorization is possible under the same
conditions of the previous section. In fact, a transformation that can
be linearized piecewise can always be performed in higher layers, on
top of other transformations, since the global group structure is not
required but weaker smoothness properties are sufficient.

\subsection{Why Hierarchical architectures: a summary}

\begin{enumerate}
\item {\it Optimization of local connections} and optimal reuse of
  computational elements.  Despite the high number of synapses on each
  neuron it would be impossible for a complex cell to pool information
  across all the simple cells needed to cover an entire image.

\item {\it Compositionality}. A hierarchical
  architecture provides signatures of larger and larger patches of the
  image in terms of lower level signatures. Because of this, it can
  access memory in a way that matches naturally with the linguistic
  ability to describe a scene as a whole and as a hierarchy of
  parts.

\item {\it Approximate factorization}. In architectures such as the
  network sketched in Fig. 1 in the main text, approximate invariance to
  transformations specific for an object class can be learned and
  computed in different stages.  This property may provide
  an advantage in terms of the sample complexity of multistage learning
  \cite{poggio03mathematics}.   For instance, approximate
  class-specific invariance to pose (e.g. for faces) can be computed
  on top of a translation-and-scale-invariant representation
  \cite{Leibo2011b}. Thus the implementation of invariance can, in some cases, be ``factorized'' into different steps corresponding to different transformations.  (see also \cite{Arathorn04,Sifre12} for related ideas).
\end{enumerate}
Probably all three properties together are the reason evolution
developed hierarchies.

\section*{4. Synopsis of Mathematical Results}
\setcounter{theorem}{0}
\noindent{\bf List of Theorems/Results}
\begin{itemize}

\item  Orbits are equivalent to probability distributions, $P_{I}$ and both are invariant and unique.\\
\noindent
{\bf Theorem}\\
The distribution $P_{I}$ is invariant and unique i.e. $I\sim I' \;\Leftrightarrow\; P_{I}=P_{I'}$.

\item  $P_{I}$ can be estimated within $\epsilon$ in terms of
    1D probability distributions of $\scal{g I}{t^{k}}$.\\
\noindent
{\bf Theorem}\\
Consider $n$ images ${\cal X}_n$ in $\cal X$. Let $K \ge
\frac{2}{c\epsilon^2} \log {\frac{n}{\delta}}$, where $c$ is a
universal constant. Then
\begin{equation*}
|d(P_I,P_{I'})
-\hat{d}_K(P_I,P_{I'})|\le\epsilon,
\end{equation*}
with probability $1 - \delta^2$, for all  $I,I'\in {\cal X}_n$.

\item Invariance from a single image based on memory of template transformations.
The simple property
\begin{equation*}
\scal{g I}{t^{k}} = \scal{I}{g^{-1}t^{k}}
\end{equation*}
implies (for unitary groups without any additional property) that the
signature components $ \mu_{n}^{k}(I)=\frac{1}{|G|} \sum_{g\in { G}}
\eta_{n}\Big(\scal{I}{gt^{k}}\Big), $ calculated on templates transformations
are invariant that is $\mu_{n}^{k}(I)=\mu_{n}^{k}(\bar{g}I)$.

\item Condition in eq. \eqref{loc} on the dot
    product between image and template implies
 invariance for Partially Observable Groups (observed
    through a window) and is equivalent to it in the case of translation and scale
    transformations.

{\bf Theorem}\\
Let $I,t\in H$ a Hilbert space, $\eta:\R\rightarrow\R^{+}$ a bijective (positive) function and $G$ a locally compact group. Let $G_{0}\subseteq G$ and suppose $supp(\scal{gI}{t})\subseteq G_{0}$. Then
\begin{eqnarray}\label{loc}
&&\scal{gI}{t^{k}}=0,\; \forall g\in G/(G_{0}\cap\bar{g}G_{0})\nonumber\\
&&\textrm{or equivalently}\;\;\;\;\;\;\;\;\;\;\;\;\;\;\;\;\;\;\;\;\;\;\;\;\;\;\;\;\Rightarrow\;\mu^{k}_{n}(I) = \mu^{k}_{n}(\bar{g}I)\nonumber \\
&&\scal{gI}{t^{k}}\neq 0,\; \forall g\in G_{0}\cap\bar{g}G_{0}
\end{eqnarray}

\item Condition in Theorem \ref{generalinv} is equivalent to
    a localization or sparsity property of the dot
    product between image and template ($\scal{I}{gt} = 0$ for
    $g\not\in G_{L}$, where $G_{L}$ is the subset of $G$ where the dot product is localized). In particular

{\bf Proposition}\\
Localization is necessary and sufficient for translation and scale
invariance. Localization for translation (respectively scale) invariance is equivalent to
the support of $t$ being small in space (respectively in frequency).

\item Optimal simultaneous invariance to translation and scale can be
  achieved by Gabor templates.

{\bf Theorem}\\
 Assume invariants are computed from
  pooling within a pooling window a set of linear filters. Then  the
  optimal templates of filters for maximum simultaneous
  invariance to translation and scale are Gabor
  functions $t(x) = e^{- \frac{x^2}{2 \sigma^2}}  e^{i \omega_0 x}$.

\item Approximate invariance can be obtained if there is approximate
    sparsity of the  image in the dictionary of templates.
  Approximate localization (defined as $\scal{t}{g t} < \delta$ for $g\not\in G_{L}$, where $\delta$ is small in the order of $\approx
  \frac{1}{\sqrt{d}}$ and $\scal{t}{g t} \approx 1$ for $g\in G_{L}$) is
  satisfied by templates (vectors of dimensionality $n$) that are
  similar to images in the set and are sufficiently ``large'' to be
  incoherent for ``small'' transformations.

\item  Approximate invariance for smooth (non group)
    transformations.

{\bf Proposition}
$\mu^{k}(I)$ is locally invariant {\bf if}
\begin{itemize}
\item $I$ is sparse in the
dictionary $t^k$;
\item $I$ and $t^k$ transform in the same way (belong to the same
  class);
\item the transformation is sufficiently smooth.
\end{itemize}

\item Sparsity of $I$ in the dictionary $t^k$ under $G$ increases with
  size of the neural images and provides invariance to clutter.

The definition is $\scal{I}{gt} < \delta$ for $g\not\in G_{L}$, where
$\delta$ is small in the order of $\approx \frac{1}{\sqrt{n}}$ and $\scal{I}{g t} \approx 1$ for $g\in G_{L}$.\\
Sparsity of $I$  in $t^k$ under $G$ improves with dimensionality of
the space $n$ and with noise-like encoding of $I$ and $t$.\\
If $n_1, n_2$ are additive uncorrelated spatial noisy clutter $\scal{I+n_1}{gt
+n_2} \approx \scal{I}{gt}$.

\item Covariance of the hierarchical architecture.\\
\noindent
{\bf Proposition}\\
The operator $\mu_{\ell}$ is covariant with respect to a non abelian (in general) group transformation, that is
$$\mu_\ell(gI) = g \mu_\ell(I).$$

\item Factorization.
{\bf Proposition}
  Invariance to separate subgroups of affine group cannot be obtained
  in a sequence of layers while factorization of the ranges of
  invariance can (because of covariance). Invariance to a smooth (non
  group) transformation can always be performed in higher layers, on
  top of other transformations, since the global group structure is
  not required.

\item Uniqueness of signature.
\small
{\bf Conjecture:}\normalsize
{\it The neural image at the first layer is uniquely represented by the final signature at the top of the hierarchy and the means and norms at
  each layer}.

\end{itemize}

\section*{5. General Remarks on the Theory}
\begin{enumerate}

\item The second regime of localization (sparsity) can be considered as a way to
  deal with situations that do not fall under the general rules (group
  transformations) by creating a series of exceptions, one for each
  object class.

\item Whereas the first regime ``predicts'' Gabor tuning of neurons in the
  first layers of sensory systems, the second regime predicts
  cells that are tuned to much more complex features, perhaps similar to neurons in inferotemporal cortex.
\item The {\it sparsity condition under the group} is related to
  properties used in associative memories for instance of the
  holographic type (see \cite{Poggio1975}). If the sparsity
  condition holds only for $I=t^k$ and for very small $a$ then it
  implies strictly memory-based recognition.
\item The theory is memory-based. It also view-based. Even assuming
  3D images (for instance by using stereo information) the various
  stages will be based on the use of 3D views and on stored sequences
  of 3D views.
\item The mathematics of the class-specific modules at the top of the
  hierarchy -- with the underlying localization condition -- justifies
  old models of viewpoint-invariant recognition (see
  \cite{PoggioEdelmann1990}).
\item The remark on factorization of general transformations implies
  that  layers dealing with general transformations can be on top of
  each other. It is possible -- as empirical results by Leibo and Li
  indicate -- that a second layer can improve the invariance to a
  specific transformation of a lower layer.
\item The theory developed here for  vision also applies to other sensory
modalities, in particular speech.
\item The theory represents a general framework for using
representations that are invariant to
transformations that are learned in an unsupervised way in order to
reduce the sample complexity of the supervised learning step.
\item Simple cells (e.g. templates) under the action of the affine group span
  a set of positions and scales and orientations. The size of their
  receptive fields therefore spans a range. The pooling window can be
  arbitrarily large -- and this does not affect selectivity when the CDF
  is used for pooling. A large pooling window implies that the
  signature is given to large patches and the signature is invariant
  to uniform affine transformations of the patches within the
  window. A hierarchy of pooling windows provides signature to patches
  and subpatches and more invariance (to more complex transformations).
\item Connections with the {\it Scattering Transform}.
\begin{itemize}
\item Our theorems about optimal invariance to scale and translation
  implying Gabor functions (first regime) may provide a justification for the use of
  Gabor wavelets by Mallat \cite{MallatGroupInvariantMain}, that does not depend
  on the specific use of the modulus as a pooling mechanism.
\item Our theory justifies several different kinds of pooling of which
  Mallat's seems to be a special case.
\item With the choice of the modulo as a pooling mechanisms, Mallat
  proves a nice property of Lipschitz continuity on
  diffeomorphisms. Such a property is not valid {\it in general} for our
  scheme where it is replaced by a hierarchical {\it parts and wholes}
  property which can be regarded as an approximation, as refined as
  desired, of the continuity w.r.t. diffeomorphisms.
\item Our second regime does not have an obvious corresponding notion in
  the scattering transform theory.

\end{itemize}

\item The theory characterizes under which conditions the signature
  provided by a HW module at some level of the hierarchy is invariant
  and therefore could be used for retrieving information (such as the
  label of the image patch) from memory. The simplest scenario is that
  signatures from modules at all levels of the hierarchy (possibly not
  the lowest ones) will be checked against the memory. Since there are
  of course many cases in which the signature will not be invariant
  (for instance when the relevant image patch is larger than the
  receptive field of the module) this scenario implies that the step
  of memory retrieval/classification is selective enough to discard
  efficiently the ``wrong'' signatures that do not have a match in
  memory. This is a nontrivial constraint. It probably implies that
  signatures at the top level should be matched first (since they are
  the most likely to be invariant and they are fewer) and lower level
  signatures will be matched next possibly constrained by the results
  of the top-level matches -- in a way similar to {\it reverse
    hierarchies} ideas. It also has interesting implications for
  appropriate encoding of signatures to make them optimally
  quasi-orthogonal e.g. incoherent, in order to minimize memory
  interference. These properties of the representation depend on
  memory constraints and will be object of a future paper on memory
  modules for recognition.

\item There is psychophysical and neurophysiological evidence that the
  brain employs such learning rules (e.g. \cite{Wallis2001, Li2008}
  and references therein). A second step of Hebbian learning may be
  responsible for wiring a complex cells to simple cells that are
  activated in close temporal contiguity and thus correspond to the
  same patch of image undergoing a transformation in time
  \cite{Foldiak1991}. Simulations show that the system could be
  remarkably robust to violations of the learning rule's assumption
  that temporally adjacent images correspond to the same object
  \cite{isik2012learning}. The same simulations also suggest that the
  theory described here is qualitatively consistent with recent
  results on plasticity of single IT neurons and with
  experimentally-induced disruptions of their invariance
  \cite{Li2008}.

\item Simple and complex units do not need to correspond to different
  cells: it is conceivable that a simple cell may be a cluster of
  synapses on a dendritic branch of a complex cell with nonlinear
  operations possibly implemented by active properties in the
  dendrites.

\item \emph{Unsupervised learning of the template orbit}.
While the templates need not be related to the test images (in the affine case), during development, the model still needs to observe the orbit of some templates. We conjectured that this could be done by unsupervised learning based on the temporal adjacency assumption \cite{Foldiak1991,wiskott2002slow}. One might ask, do ``errors of temporal association'' happen all the time over the course of normal vision? Lights turn on and off, objects are occluded, you blink your eyes -- all of these should cause errors. If temporal association is really the method by which all the images of the template orbits are associated with one another, why doesn't the fact that its assumptions are so often violated lead to huge errors in invariance?

The full orbit is needed, at least in theory. In practice we have found that significant scrambling is possible as long as the errors are not correlated. That is, normally an HW-module would pool all the $\scal{I}{g_i t^k}$. We tested the effect of, for some $i$, replacing $t^k$ with a different template $t^{k^\prime}$. Even scrambling $50\%$ of our model's connections in this manner only yielded very small effects on performance. These experiments were described in more detail in \cite{isik2012learning} for the case of translation. In that paper we modeled Li and DiCarlo's "invariance disruption" experiments in which they showed that a temporal association paradigm can induce individual IT neurons to change their stimulus preferences under specific transformation conditions \cite{Li2008,Li2010}. We also report similar results on another "non-uniform template orbit sampling" experiment with 3D rotation-in-depth of faces in \cite{Liao2013}.
\end{enumerate}

\section*{6. Empirical support for the theory}

The theory presented here was inspired by a set of related computational models for visual recognition, dating from 1980 to the present day.  While differing in many details, HMAX, Convolutional Networks \cite{lecun1995convolutional}, and related models use similar structural mechanisms to hierarchically compute \emph{translation} (and sometimes \emph{scale}) invariant signatures for progressively larger pieces of an input image, completely in accordance with the present theory.

With the theory in hand, and the deeper understanding of invariance it provides, we have now begun to develop a new generation of models that incorporate invariance to larger classes of transformations.

\subsection{Existing models}

\begin{figure}\centering
\includegraphics[width=0.4\textwidth]{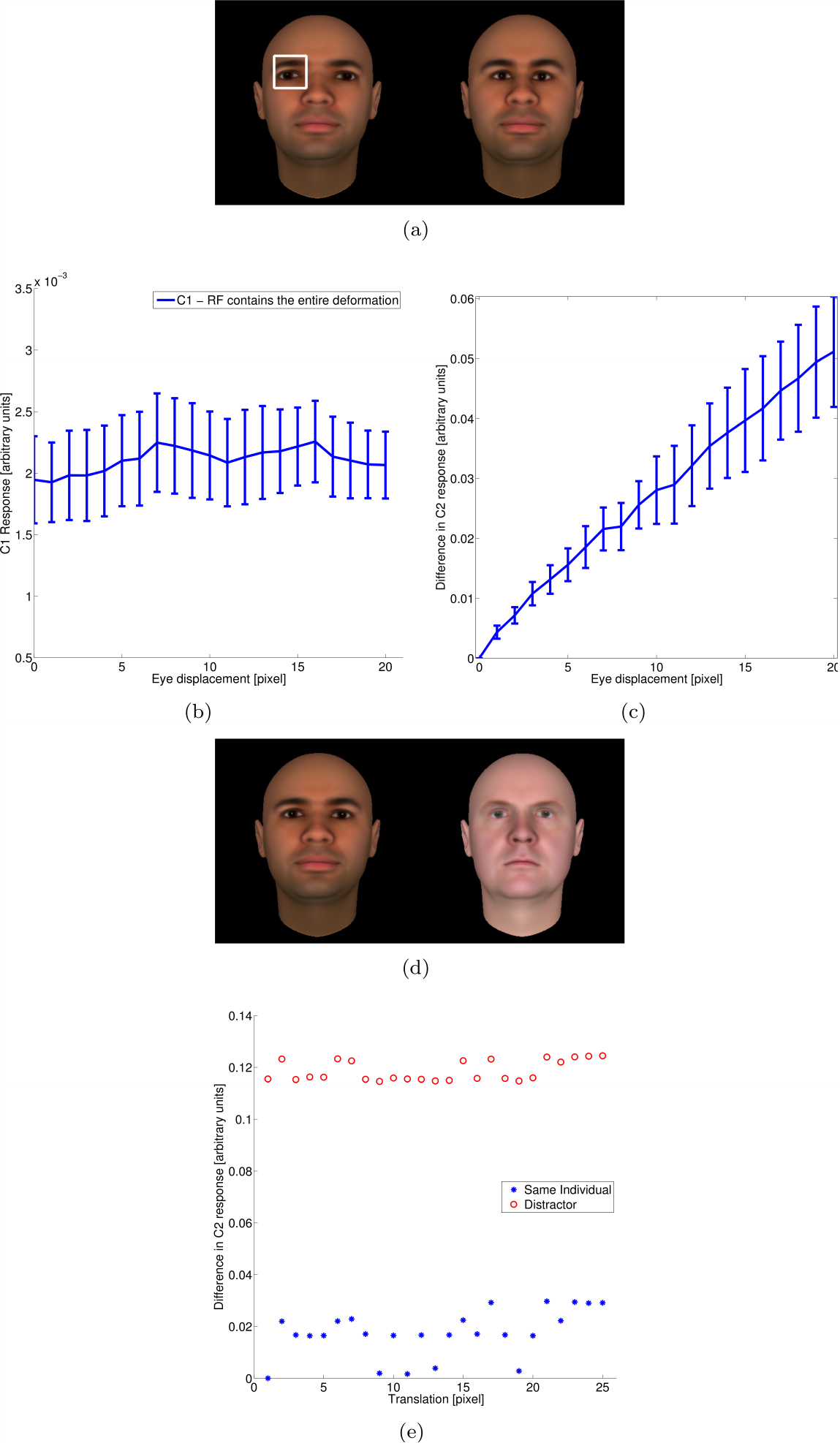}
\caption{Empirical demonstration of the properties of invariance, stability and uniqueness of the hierarchical
    architecture in a specific 2 layers implementation (HMAX). Inset (a) shows the reference image on the left and a
    deformation of it (the eyes are closer to each other) on the
    right; (b) shows that an HW-module in layer 1 whose
    receptive fields covers the left eye provides a signature vector  ($C_1$)  which is invariant to the
    deformation; in (c) an HW-module at layer 2 ($C_2$) whose receptive fields contain the
    whole face provides a signature vector which is (Lipschitz) stable with respect to the
    deformation. In all cases, the Figure shows just the Euclidean norm of the signature vector. Notice that the $C_1$ and $C_2$ vectors are not only
    invariant but also selective. Error bars represent $\pm1$ standard deviation.
    Two different images (d)
    are presented at various location in the visual field. The Euclidean distance between the signatures of a set of HW-modules at layer 2  with the same
    receptive field (the whole image) and a reference vector is shown in (e).
    The signature vector is invariant to global
    translation and discriminative (between the two faces). In this
    example the HW-module represents the top of a hierarchical,
    convolutional architecture. The images we used
    were 200$\times$200 pixels\label{invariancestability2}}
\end{figure}

Fukushima's Neocognitron \cite{Fukushima1980} was the first of a class of recognition models consisting of hierarchically stacked modules of simple and complex cells (a ``convolutional'' architecture).  This class has grown to include Convolutional Networks, HMAX, and others \cite{Pinto2009a,Yamins2013}.  Many of the best performing models in computer vision are instances of this class.  For scene classification with thousands of labeled examples, the best performing models are currently Convolutional Networks \cite{NIPS2012HintonImagenet}.  A variant of HMAX \cite{Mutch2006} scores 74\% on the Caltech 101 dataset, competitive with the state-of-the-art for a single feature type.  Another HMAX variant added a time dimension for action recognition \cite{jhuang2010}, outperforming both human annotators and a state-of-the-art commercial system on a mouse behavioral phenotyping task.  An HMAX model \cite{Serre2007} was also shown to account for human performance in rapid scene categorization.  A simple illustrative empirical demonstration of the HMAX properties of invariance, stability and uniqueness is in figure \ref{invariancestability2}.

All of these models work very similarly once they have been trained.  They all have a convolutional architecture and compute a high-dimensional signature for an image in a single bottom-up pass.  At each level, complex cells pool over sets of simple cells which have the same weights but are centered at different positions (and for HMAX, also scales).  In the language of the present theory, for these models, $g$ is the 2D set of translations in $x$ and $y$ (3D if scaling is included), and complex cells pool over partial orbits of this group, typically outputting a single moment of the distribution, usually sum or max.

The biggest difference among these models lies in the training phase.  The complex cells are fixed, always pooling only over position (and scale), but the simple cells learn their weights (templates) in a number of different ways.  Some models assume the first level weights are Gabor filters, mimicking cortical area V1.  Weights can also be learned via backpropagation, via sampling from training images, or even by generating random numbers.  Common to all these models is the notion of automatic \emph{weight sharing}: at each level $i$ of the hierarchy, the $N_i$ simple cells centered at any given position (and scale) have the same set of $N_i$ weight vectors as do the $N_i$ simple cells for every other position (and scale).  Weight sharing occurs by construction, not by learning, however, the resulting model is equivalent to one that learned by observing $N_i$ different objects translating (and scaling) everywhere in the visual field.

One of the observations that inspired our theory is that in convolutional architectures, random features can often perform nearly as well as features learned from objects \cite{leibo2010learning,ICML2011Saxe,Jarrett2009,Yamins2013} -- the architecture often matters more than the particular features computed.  We postulated that this was due to the paramount importance of invariance.  In convolutional architectures, invariance to translation and scaling is a property of the architecture itself, and objects in images always transform and scale in the same way.

\subsection{New models}

Using the principles of invariant recognition made explicit by the present theory, we have begun to develop models that incorporate invariance to more complex transformations which, unlike translation and scaling, cannot be solved by the architecture of the network, but must be learned from examples of objects undergoing transformations.  Two examples are listed here.

{\bf Faces rotating in 3D.} In \cite{Leibo2011b}, we added a third H-W layer to an existing HMAX model which was already invariant to translation and scaling.  This third layer modeled invariance to rotation in depth for faces.  Rotation in depth is a difficult transformation due to self-occlusion.  Invariance to it cannot be derived from network architecture, nor can it be learned generically for all objects.  Faces are an important class for which specialized areas are known to exist in higher regions of the ventral stream.  We showed that by pooling over stored views of template faces undergoing this transformation, we can recognize novel faces from a single example view, robustly to rotations in depth.

{\bf Faces undergoing unconstrained transformations.}  Another model \cite{Liao2013} inspired by the present theory recently advanced the state-of-the-art on the Labeled Faces in the Wild dataset, a challenging same-person / different-person task.  Starting this time with a first layer of HOG features \cite{Dalal2005}, the second layer of this model built invariance to translation, scaling, and limited in-plane rotation, leaving the third layer to pool over variability induced by other transformations.  Performance results for this model are shown in figure \ref{LFW} in the main text.

\end{article}

\end{document}